\documentclass{article}
\usepackage[sort&compress,numbers]{natbib}
\usepackage{graphicx} 
\usepackage[utf8]{inputenc} 
\usepackage[T1]{fontenc}    
\usepackage{hyperref} 
\hypersetup{
  colorlinks=false,
  pdfborder={0 0 0}
}
\usepackage{url}            
\usepackage{booktabs}       
\usepackage{amsfonts}       
\usepackage{nicefrac}       
\usepackage{microtype}      
\usepackage{xcolor} 
\usepackage{multirow}
\usepackage{multicol}
\usepackage{amsmath}
\usepackage{authblk}
\usepackage{float}
\usepackage{listings}

\usepackage{etoolbox}
\patchcmd{\abstract}{\vspace*{\baselineskip}}{\vspace{0.5\baselineskip}}{}{}
\date{{\small 30th March 2026}}

\title{\textit{Mimosa} Framework: Toward Evolving Multi-Agent Systems for Scientific Research}

\newcommand{\cocorresp}{\textsuperscript{*}}

\author[1,2]{Martin Legrand\cocorresp}
\author[1,2]{Tao Jiang\cocorresp}
\author[1,2]{Matthieu Feraud}
\author[1, 2, 3]{Benjamin Navet}
\author[1, 2, 3]{Yousouf Taghzouti}
\author[2, 3]{Fabien Gandon}
\author[1]{Elise Dumont}
\author[1,2]{Louis-F\'elix Nothias\cocorresp}

\affil[1]{Universit\'e C\^ote d'Azur, CNRS, ICN, Nice, France}
\affil[2]{Interdisciplinary Institute for Artificial Intelligence (3iA) C\^ote d'Azur, Sophia-Antipolis, France}
\affil[3]{Inria, Université Côte d'Azur, CNRS, I3S, France}

\date{30th March 2026}

\begin{document}

\maketitle
\begingroup
\renewcommand{\thefootnote}{}
\footnotetext{*Co-corresponding authors: martin.legrand@univ-cotedazur.fr, tao.jiang@univ-cotedazur.fr, louis-felix.nothias@univ-cotedazur.fr}
\endgroup
\setcounter{footnote}{0}

\begin{abstract}
Current Autonomous Scientific Research (ASR) systems, despite leveraging
large language models (LLMs) and agentic architectures, remain constrained by
fixed workflows and toolsets that prevent adaptation to evolving tasks
and environments. We introduce \textit{Mimosa}, an evolving multi-agent
framework that automatically synthesizes task-specific multi-agent workflows
and iteratively refines them through experimental feedback.

\textit{Mimosa} leverages the Model Context Protocol (MCP) for dynamic tool
discovery, generates workflow topologies via a \textit{meta-orchestrator}, executes
subtasks through code-generating agents that invoke available tools and
scientific software libraries, and scores executions with an LLM-based \textit{judge}
whose feedback drives workflow refinement.

On \textit{ScienceAgentBench}, \textit{Mimosa} achieves a success rate of 43.1\%
with DeepSeek-V3.2, surpassing both single-agent baselines and static
multi-agent configurations. Our results further reveal that models respond
heterogeneously to multi-agent decomposition and iterative learning,
indicating that the benefits of workflow evolution depend on the
capabilities of the underlying execution model.

Beyond these benchmarks, \textit{Mimosa}'s modular architecture and
tool-agnostic design make it readily extensible, and its fully logged
execution traces and archived workflows support auditability by preserving
every analytical step for inspection and potential replication. Combined
with domain-expert guidance, the framework has the potential to automate a
broad range of computationally accessible scientific tasks across
disciplines. Released as a fully open-source platform, \textit{Mimosa}
aims to provide an open foundation for community-driven ASR.
\end{abstract}

\newpage
\section{Introduction}

\paragraph{Context and Problem.}

Advances in automation, high-throughput experimentation, and computation
have greatly expanded the scientific community's capacity to generate
data---in fields such as biology, drug discovery, materials science, and
climate modeling---but the ability to transform these data into actionable
knowledge has not kept pace~\cite{Campo2022,Cambrosio2020}. Researchers must
continually arbitrate between breadth and depth of analysis under the
practical constraints of time, expertise, and computational
resources~\cite{Berger2023}, even as pressing societal challenges
underscore the need for methods that accelerate knowledge generation
beyond current limits~\cite{Gil2014}.

Compounding this bottleneck, the fragmentation of methods, tools, and
reporting standards across laboratories has produced a reproducibility
crisis in which critical findings become difficult or impossible to
verify~\cite{Baker2016}. Reliable and unreliable results intermingle
within the scientific record, growing increasingly
indistinguishable~\cite{Ioannidis2005}. When subsequent research builds
on this uncertain foundation without efficient means to audit prior
claims, the risk of misdirecting future work
escalates~\cite{Beaulieu-Jones2017}.

Addressing these interconnected challenges requires systems that not only
automate scientific tasks at scale---whether purely computational or
involving laboratory instruments through digital interfaces---but do so
through transparent, adaptable workflows whose every analytical step is
recorded and auditable, both during the production cycle and when results
are communicated~\cite{Beaulieu-Jones2017}.

\paragraph{State of the Art and Gap.}

Emerging paradigms such as Autonomous Scientific Research (ASR) and
Self-Driving Labs (SDLs)~\cite{aag2024chemrev,luo-jacs25} offer a
promising path forward by automating core scientific tasks and
experimentation. Recent systems leverage large language models (LLMs)
with expanded context windows, domain-specialized agentic architectures,
and reasoning frameworks such as tree-of-thought and tool-augmented
agents that execute multi-step experimental protocols with increasing
sophistication~\cite{8_huang_towards_2023,1_tang_ai-researcher_2025}.
State-of-the-art platforms propose to integrate instrument control, literature
retrieval, and iterative hypothesis refinement within unified
pipelines~\cite{zhou2025autonomousagentsscientificdiscovery}.

However, current AI-agent systems exhibit two interconnected limitations
when deployed in complex, real-world scientific settings:
(1)~limitations in long-horizon LLM-based execution, including context
window constraints and degraded use of relevant information across
extended trajectories, which are associated with failure modes such as
information loss, hallucinations, attention dilution, and semantic
drift~\cite{du2025contextlengthhurtsllm,liu2024lostmiddle}; and
(2)~architectural rigidity, namely a limited ability to reorganize agent
coordination, communication structure, and tool use in response to
evolving experimental demands, unexpected tool failures, or non-linear
discovery trajectories~\cite{zhou2025autonomousagentsscientificdiscovery}.

In practice, these systems typically rely on fixed toolsets, computational libraries, and
predefined coordination protocols, which prevent them from
reconfiguring when new instruments or software resources are introduced, objectives are
revised, or intermediate results call for alternative analytical paths.
Consequently, even sophisticated systems remain bounded by their initial
configuration and function as powerful but brittle automation tools
rather than genuinely adaptive scientific systems.

\paragraph{Research Questions and Contributions.}

Our central research question is: \emph{How can multi-agent systems
adjust their coordination and tool use as scientific tasks change,
without losing important context or becoming too rigid to adapt?} A
complementary question asks how such a system can remain capable across
the full breadth of tasks that scientific research demands---from data
acquisition and computational analysis to control of experimental
apparatus through programmatic interfaces, result interpretation, and
knowledge synthesis---rather than excelling only in narrow, predefined
operations. A third, longer-term motivation concerns the computational
reproducibility of published scientific results.

To make this concrete, consider computational drug design. A typical
study runs through several connected stages: virtual screening, molecular
docking, and molecular dynamics simulations~\cite{Sliwoski2014}. Each stage requires different
expertise and different decisions, often under competing scientific and
computational constraints. Results from later stages frequently challenge
earlier assumptions, requiring iterative revision of parameters,
hypotheses, or analytical paths~\cite{Sadybekov2023}. Such recursive
adjustment is constitutive of complex scientific inquiry, particularly in
interdisciplinary and computationally intensive settings. Current AI
systems, however, lack robust mechanisms to maintain specialized roles
across stages, critically reassess earlier decisions, or adapt focus as
new findings emerge. Single-agent systems and fixed pipelines offer no
reliable means of doing so, and the same rigidity limits the potential
for revisiting prior computational analyses in a systematic way.

We address these challenges through \textit{Mimosa}, an open-source
framework for automated computational scientific tasks. Its modular
architecture leverages the Model Context
Protocol (MCP)~\cite{anthropic2024mcp} for dynamic integration of diverse
computational tools and laboratory instruments. Rather than relying on
static pipelines, \textit{Mimosa} dynamically composes task-specific
workflows and refines them in response to execution feedback.

This adaptability is sustained by a meta-learning layer that
continuously refines how agentic workflows are designed and coordinated
based on prior performance.

\textit{Mimosa} offers an open alternative to closed, rigid systems for
automating time-consuming scientific tasks from minimal task
instructions. More broadly, it is designed to help researchers scale
analytical work more effectively and to support efforts toward more
reliable and auditable scientific discovery.

\paragraph{Code availability.}

Both \textit{Mimosa} and \textit{Toolomics}, our companion platform for MCP server management, are released as open-source software under the Apache License 2.0, a permissive open-source license enabling broad reuse, modification, and integration across research and industrial environments:

\begin{center}
  \url{https://github.com/HolobiomicsLab/Mimosa-AI}
\end{center}

\begin{center}
  \url{https://github.com/HolobiomicsLab/Toolomics}
\end{center}

\section{Related work}

\subsection{Autonomous Scientific Research (ASR)}

The emergence of LLMs has catalyzed a paradigm shift from ``AI for
Science'', where models act as static predictors, to ``agentic science'',
characterized by autonomous decision-making and iterative
reasoning~\cite{2_wei_ai_2025}. Unlike traditional automation restricted
to predefined tasks, ASR leverages the reasoning and planning capabilities
of agentic systems to operate across multiple stages of the research
lifecycle~\cite{zhou2025autonomousagentsscientificdiscovery}, with recent
surveys noting a move beyond literature synthesis toward more active roles
in hypothesis generation and experimental
design~\cite{3_zhang_evolving_2025}. A critical frontier now shifts
attention from model capacity alone to the scalability of agentic systems:
their ability to manage complex tasks, large scientific datasets, and
multiple concurrent workflows with reasonable computational and economic
efficiency~\cite{1_tang_ai-researcher_2025}. In this regime, bottlenecks
arise from the rigidity of agent
coordination~\cite{21_kim_towards_2025}, saturation of context windows
with data returned by scientific
tools~\cite{labate2025solvingcontextwindowoverflow}, and brittle
decision-making over extended discovery
trajectories~\cite{wang2026reasoningfailsplanplanningcentric}.

In domains such as molecular docking and materials discovery, ASR systems
provide a dual benefit: they leverage domain-specific expertise and
state-of-the-art cheminformatics tools to explore chemical spaces far
beyond human capacity, while making the scientific process inherently
reproducible and auditable by
design~\cite{1_tang_ai-researcher_2025,2_wei_ai_2025}. Closed-loop
experimentation is particularly transformative in materials science and
catalysis, where experimental landscapes are often non-linear and highly
complex~\cite{aag2024chemrev}. By offloading execution and
optimization to autonomous platforms, researchers can prioritize
high-level analysis, experimental strategy, theory development and interpretation.

While ASR focuses on the cognitive and computational aspects of discovery,
Self-Driving Labs (SDLs) represent the physical realization of these
principles through the integration of AI-driven decision-making with
programmatically accessible laboratory
instruments~\cite{aag2024chemrev}. SDLs close the experimental
loop by enabling AI systems to analyze outputs, select informative next
experiments, and iteratively adapt experimental
design~\cite{5_tobias_autonomous_2025}. As these systems increasingly
close the loop, the primary limitation shifts from task-specific
automation to the design of adaptive agentic architectures capable of
sustaining long-horizon reasoning, coordination, and methodological
self-revision under sparse and delayed scientific
feedback~\cite{zhou2025autonomousagentsscientificdiscovery}. These challenges motivate a closer examination of the agentic architectures that underpin such systems.

\subsection{Agentic Architectures}

The transition from static LLMs to agentic systems is defined by the integration of reasoning, tool-use, agent-to-agent collaboration and iterative planning \cite{8_huang_towards_2023}. 
While much of this work targets coding, mathematics, and general reasoning benchmarks, the architectural patterns it introduces---task decomposition, role specialization, and automated agent design---are directly relevant to scientific computing and discovery.
Agents have demonstrated initial success in scientific domains such as chemical synthesis planning \cite{12_xuan-vu_synthelite_2025} and mechanistic enzyme design \cite{16_jacob_beyond_2025}, reflecting increasingly active engagement with AI-driven scientific design in these areas.

Long-horizon adaptability further requires mechanisms for strategy revision, self-reflection, and coordination updates in response to empirical feedback \cite{zhou2025autonomousagentsscientificdiscovery}. Multiple paradigms in architecture have emerged to address these issues.

\subsubsection{Single-Agent Systems}

Early frameworks rely on a single tool-augmented agent to handle multiple stages of a given research task. 
While capable of accelerating hypothesis generation in specialized fields like protein design \cite{16_jacob_beyond_2025}, single-agent systems are prone to performance collapse when faced with unfamiliar tasks such as newest Kaggle challenges \cite{13_huang_mlagentbench_2024} or novel research challenges \cite{14_zhang_mlrc-bench_2025}.

A critical failure mode is the inability to recover from early erroneous hypotheses, as a single reasoning trajectory offers no mechanism to escape a flawed line of inquiry once committed, leading to terminal reasoning loops. Without external critique or adaptive strategy revision, early sampling errors can irreversibly corrupt the entire discovery trajectory \cite{zhou2025autonomousagentsscientificdiscovery}.

\subsubsection{Multi-Agent Systems (MAS)}

To address the limitations of single-agent systems, MAS frameworks distribute cognitive load across specialized roles---experimenters, analysts, reviewers---sometimes equipped with dedicated tools, collaborating to verify results \cite{20_team_internagent_2025}.  

While multi-agent coordination has been shown to degrade performance
on sequential reasoning tasks~\cite{21_kim_towards_2025}, other work
establishes that under specific prompting conditions, MAS can develop
measurable emergent coordination that improves task performance through
complementary role
specialization~\cite{riedl2025emergentcoordinationmultiagentlanguage}.

Furthermore, recent studies suggest mainstream MAS operate as constrained finite-state machines with predefined transition protocols~\cite{zhang2025metaagentautomaticallyconstructingmultiagent}, where fixed coordination protocols conflict with the open-ended and non-linear nature of real-world scientific inquiry~\cite{anthropic2025multiagentresearch}.

Scientific discovery requires agents to dynamically reconfigure their interaction topology in response to task complexity, uncertainty, and unexpected experimental outcomes, capabilities largely absent from static MAS designs \cite{2_wei_ai_2025}.

\subsubsection{Automated Agent Design and Self-Evolving Systems}

A prominent line of work treats the design of agentic systems itself as a search problem. ADAS \cite{23_hu_automated_2025} introduces a meta-agent that iteratively programs new agents in an unconstrained code space, maintaining an archive of discovered designs. AgentSquare \cite{shang2024agentsquare} proposes a modular design space decomposing agents into four functional modules (planning, reasoning, tool use, memory) and searches over module combinations via evolution and recombination. GPTSwarm \cite{zhuge2024gptswarm} models multi-agent systems as computational graphs and optimizes edge weights and node operations. EvoAgent \cite{yuan2024evoagent} uses evolutionary algorithms to automatically extend single-agent systems into multi-agent populations. At the prompt level, Promptbreeder \cite{fernando2024promptbreeder} and EvoPrompt~\cite{guo2024evoprompt} demonstrate that prompts can be evolved through iterative mutation and selection. Many of these systems were introduced and evaluated on coding, mathematics, and general reasoning tasks, though some have also been tested in broader tool-use settings.

At the frontier of this design-as-search paradigm, self-evolving systems exhibit meta-learning --- the ability to autonomously modify their own code, tools, and architectures \cite{23_hu_automated_2025,26_sheng_self-programming_2023}. For scientific agents, optimization and evolution are not merely about refining parameters, but about improving how agents navigate the scientific landscape --- from leveraging prior knowledge and reasoning strategies to mastering specialized tools and structuring memory~\cite{zhou2025autonomousagentsscientificdiscovery}. Systems such as the Darwin Gödel Machine \cite{22_zhang_darwin_2025} and MAS-Zero \cite{24_ke_mas-zero_2025} explore open-ended self-modification, while CASCADE \cite{28_huang_cascade_2025} extends cumulative executable skill acquisition to scientific domains.

However, although much of this literature was initially validated on coding and reasoning benchmarks, recent work has begun extending self-improving agent design to scientific computing tasks. These tasks place greater emphasis on domain-specific tools, heterogeneous data formats, and experimentally or domain-grounded outputs than the benchmark settings that motivated much of the early literature. The remaining gap reflects both the relatively recent emergence of suitable scientific benchmarks and a difference in search space: many of these systems evolve unconstrained agent code, agent-level strategies, or coordination patterns, whereas scientific workflows require explicit organization around heterogeneous tools, structured task decomposition, and domain-grounded validation. Direct comparison on a shared benchmark remains an open challenge.

Recent automated multi-agent architecture search methods broaden this design space in complementary directions. MaAS~\cite{24_ke_mas-zero_2025} replaces the search for a single static architecture with an agentic supernet from which query-dependent workflows are sampled at inference time. AutoMaAS~\cite{zhang2025metaagentautomaticallyconstructingmultiagent} extends this line with neural architecture search-inspired operator generation, fusion, and elimination under joint performance--cost objectives. MetaAgent~\cite{zhang2025metaagentautomaticallyconstructingmultiagent} casts system construction as finite-state-machine synthesis with explicit state transitions, while MAS-Zero~\cite{24_ke_mas-zero_2025} performs inference-time design--critique--refine without a held-out validation set.

Recent ASR systems illustrate how workflow-design choices translate into end-to-end scientific capability. The AI Scientist-v2 advances end-to-end scientific paper production through agentic tree search~\cite{yamada2025aiscientistv2}, while CASCADE~\cite{28_huang_cascade_2025} introduces both a cumulative skill-acquisition framework and its companion SciSkillBench benchmark for chemistry and materials tasks. On the evaluation side, BixBench~\cite{mitchener2025bixbench} provides a dedicated benchmark for LLM-based agents in computational biology, and ScienceAgentBench~\cite{33_chen2025scienceagentbenchrigorousassessmentlanguage} offers 102 data-driven discovery tasks spanning four scientific disciplines. Together these developments underscore rapid progress in ASR while leaving open the question of how to achieve auditable, adaptive workflow evolution across heterogeneous scientific tasks. 
\subsection{Summary of open challenges}

Across these paradigms, four interconnected limitations persist. 
First, single-agent frameworks suffer from semantic drift, reasoning collapse, 
and progressive loss of \textit{epistemic memory} as context 
accumulates~\cite{liu2024lostmiddle,arike2025goaldrift,30_zheng_lifelong_2026}, 
leading to terminal reasoning loops and repetitive errors~\cite{14_zhang_mlrc-bench_2025,10_gu_blade_2025}.

Second, multi-agent systems mitigate drift through specialization
but remain bound by static coordination protocols that cannot adapt
when experiments behave unexpectedly, new tools are introduced, or
existing tool documentation is updated~\cite{2_wei_ai_2025}.

Third, self-evolving architectures have demonstrated meta-learning
on coding and reasoning benchmarks~\cite{23_hu_automated_2025,22_zhang_darwin_2025},
and recent work has begun extending cumulative skill acquisition to
scientific domains, notably chemistry and materials
science~\cite{28_huang_cascade_2025}. In parallel, a growing number
of systems target heterogeneous scientific tool ecosystems in
biomedicine and bioinformatics, with dedicated benchmarks also
emerging~\cite{mitchener2025bixbench}. However, these capabilities are
still rarely demonstrated together: cumulative skill growth, broad
scientific domain coverage, and architecture-level self-modification
typically remain distributed across separate lines of work and
validation settings.

Fourth, existing ASR systems still seldom treat the \textit{execution
environment} as a first-class design concern. In many systems, tool
invocations remain effectively stateless, environment configurations
are implicit, and execution traces are not captured in a form
structured for downstream verification~\cite{Beaulieu-Jones2017}. Dependence on proprietary,
closed-weight models or opaque agentic systems further compounds
this problem, creating barriers to reproducibility, auditing, and
community adoption~\cite{Baker2016}.

These gaps motivate an architecture that treats the system's own
coordination and exploration logic as a mutable, learnable component,
while embedding the reproducibility and traceability principles
established by scientific workflow managers into AI-driven research
automation.

\subsection{Positioning of \textit{Mimosa}}

Relative to the paradigms surveyed above, \textit{Mimosa} addresses
these limitations through a combination of design choices that span
workflow evolution, tool integration, and execution infrastructure.
Unlike prompt-level methods such as
Promptbreeder~\cite{fernando2024promptbreeder} and
EvoPrompt~\cite{guo2024evoprompt}, and unlike architecture search
systems such as MaAS, AutoMaAS, MetaAgent, and MAS-Zero that were
primarily introduced on general reasoning workloads, \textit{Mimosa}
targets scientific computing and combines four design choices that are
central to its evolutionary design and still uncommon in combination:
(1)~it operates over DAG-structured workflows with explicit agent
roles, communication edges, and tool allocations rather than
unconstrained code or fixed module slots; (2)~it integrates dynamic
tool discovery via MCP, enabling evolved workflows to adapt to changing
computational environments; (3)~it is developed for scientific research tasks and evaluated on \textit{ScienceAgentBench}~\cite{33_chen2025scienceagentbenchrigorousassessmentlanguage}, where ground-truth assessment is available, providing both a fitness
signal for workflow evolution and an empirical testbed that remains
comparatively uncommon in prior automated design work; and
(4)~it deploys tools as containerised MCP servers within isolated
execution environments, producing structured execution traces that
are reproducible and auditable---properties that, combined with its
open-source design, make discovered workflows verifiable and portable
across computational contexts. While \textit{Mimosa} is inspired by evolutionary design principles developed for open-ended objectives, in this work we focus on iterative refinement of task-specific workflows.

\section{Methods}
\label{sec:methods}

\subsection{Motivation for evolving workflow topology}

\subsubsection{From Single-Agent Limits to Workflow Topology}

Large language models generate outputs by sampling from a policy
$\pi_\theta(\cdot \mid p, o)$ conditioned on a prompt $p$ and tool set
$o \subseteq \mathcal{O}$, where $\theta$ are frozen weights. When equipped
with execution capabilities, an agent follows an iterative trajectory:

$$\tau = \bigl(s^{(0)},\, a^{(0)},\, r^{(0)},\; s^{(1)},\, a^{(1)},\,
r^{(1)},\; \ldots,\; s^{(n)}\bigr)$$

where $s^{(k)} \sim \pi_\theta(\cdot \mid p, \tau_{<k})$ is the model output
at step $k$, $a^{(k)} = \text{parse}(s^{(k)})$ is the extracted action or
code, and $r^{(k)} = \text{exec}(a^{(k)})$ is the result of its execution.
The agent terminates when $a^{(n)}$ invokes a terminal action, yielding
final output $s^*$.

\textbf{Semantic drift.} Context accumulates linearly across steps:

$$c_k = [p,\, s^{(0)},\ a^{(0)},\ r^{(0)},\, \ldots,\, s^{(k)}]$$ 

As $|c_k|$ grows, attention dilutes over earlier tokens and the effective optimization target shifts from task completion toward self-consistency with prior
outputs~\cite{mohsin2026fundamentallimitsllmsscale}.
This is consistent with evidence that information placed in the middle of a long context is used less reliably than information near the boundaries~\cite{liu2024lostmiddle}, and recent work shows that long-horizon agents can progressively deviate from their initial objective, a failure mode termed \emph{goal drift}~\cite{arike2025goaldrift}.
Herein, we refer to this progressive divergence as \textit{semantic drift}.

\subsubsection{Workflow Topology as a Structural Mitigation}

Multi-agent workflows address semantic drift through task decomposition.
Rather than accumulating context over a long trajectory, the problem is
decomposed into subtasks, each handled by an agent with bounded context.

\textbf{Definition.} A workflow $W$ is a directed acyclic graph:
$$W = (A,\, E)$$
where $A = \{a_1, \ldots, a_k\}$ is a set of agents (vertices) and
$E \subseteq A \times A$ defines information flow (edges). Each agent
$a_i = (p_i, o_i, \theta_i)$ is assigned a subtask $t_i$, a prompt $p_i$,
and a tool subset $o_i$, and operates with fixed model parameters $\theta_i$.
Execution proceeds by topological order: each agent receives a bounded
context $[p_i,\, s^*_{i-1}]$\footnote{This notation is a simplification for
clarity. In Mimosa's implementation, agents receive a sliding
window of recent outputs from a shared cumulative state,
permitting access beyond the immediate predecessor.} consisting of its prompt and the predecessor's
output, capping context at $O(L_\text{subtask})$ rather than
$O(k \cdot L_\text{subtask})$ for a single agent.

\paragraph{Sensitivity to Prompt Specification}
Workflow performance is sensitive to how the global task $t^G$ is specified.
An ambiguous or misaligned prompt propagates errors across the agent graph,
and no per-agent optimization recovers a workflow that is structurally
misaligned with the true objective. This motivates treating the workflow
itself as a discovery problem, iteratively reshaping topology, prompts,
and tool allocations in response to execution feedback.

\paragraph{Workflow Execution Trace}
Executing workflow $W$ yields an ordered tuple of final agent outputs:
$$\mathbf{s}^*_W = \left(s_1^*,\, s_2^*,\, \ldots,\, s_{|A|}^*\right)$$
where $s_i^*$ is the terminal output of agent $a_i \in A$, ordered by the 
topological sort of $W$.

\paragraph{Workflow Discovery Problem}
We formulate workflow discovery as local search over the discrete space of
workflow topologies $\mathcal{W}$. The objective is to
find:
\begin{equation}
    W^* = \arg\max_{W \in \mathcal{W}} J(\mathbf{s}^*_W)
\end{equation}
where $J(\mathbf{s}^*_W)$ is the expected performance of the workflow on global task
$t^G$, evaluated by an LLM-as-a-judge~\cite{zheng2023judging} over four
evaluation criteria:
\begin{equation}
    J(\mathbf{s}^*_W) = \frac{1}{4}\bigl(g(\mathbf{s}^*_W) + 
    c(\mathbf{s}^*_W) + q(\mathbf{s}^*_W) + a(\mathbf{s}^*_W)\bigr)
\end{equation}
where $g$, $c$, $q$, $a$ denote goal alignment, collaboration efficiency,
output quality, and answer plausibility respectively, each scored over the
execution trajectory $\tau$. This \textit{judge} provides approximate rather than
exact quality signal; its directional accuracy is sufficient to drive
convergence, and its correlation with benchmark ground-truth metrics is
left to future work.

Workflow mutations are single structural edits drawn from three operations:
\textit{prompt refinement} (rewriting $p_i$ for one agent),
\textit{agent addition or removal} (inserting or deleting a node from $A$
with reconnected edges), and \textit{edge rewiring} (modifying $E$ to
redirect information flow). Each mutation is proposed by an LLM given the
current workflow definition and its evaluation feedback.

\paragraph{Initialization}
For a new task $t_{\text{new}}$, we query the workflow archive $\mathcal{A}$
for prior candidates with similar task specifications:
\begin{equation}
    \mathcal{C} = \{W \in \mathcal{A} \mid \text{sim}(t_W,\, t_{\text{new}})
    > \epsilon\}
\end{equation}
where $sim$ is the embedding cosine similarity function, $\epsilon = 0.7$ is the similarity threshold and $t_W$ is the task prompt associated with workflow $W$. If $\mathcal{C} \neq \emptyset$, we
select the highest-scoring candidate $W_0 = \arg\max_{W \in \mathcal{C}}
\text{score}(W)$ where $\text{score}(W)$ is the archived performance of the workflow on $t_W$ and apply a single mutation to adapt it to $t_{\text{new}}$.
If $\mathcal{C} = \emptyset$, $W_0$ is synthesized de novo via
one-shot LLM generation from $t_{\text{new}}$ alone.
(In practice, this de novo path was taken for all evaluated tasks;
see Section~\ref{sec:limitations}.)

\paragraph{Iterative Refinement}
Starting from $W_0$, the \textit{meta-orchestrator} performs single-incumbent local
search. At each iteration $n$:
\begin{enumerate}
    \item Let $W_n$ be the best-performing workflow observed so far.
    \item Generate a neighbor $W' \leftarrow \text{Mutate}(W_n)$ via a
          single structural or prompt-level edit proposed by the LLM.
    \item Execute $W'$, collect the final output of all the agents $\mathbf{s}^*_W{}'$, and compute $J(\mathbf{s}^*_W{}')$.
    \item Accept: $W_{n+1} \leftarrow W'$ if $J(\mathbf{s}^*_W{}') > J(\mathbf{s}^*_{W_n})$,
          else $W_{n+1} \leftarrow W_n$.
\end{enumerate}
Search terminates after a predefined number of iterations or when $J(\tau) > 0.9$.
Unlike population-based evolutionary methods, only the single incumbent
generates offspring; exploration is strictly local. Through successive
refinement, the \textit{meta-orchestrator} relocates, merges, or subdivides agent
responsibilities, progressively aligning the workflow topology with the
underlying structure of the task's solution space.

\subsection{Architecture}

The \textit{Mimosa} framework is implemented as an evolving MAS, decomposed into five distinct layers: (0)~an optional \textit{planning} layer that decomposes high-level scientific goals into individual tasks; (1)~a \textit{tool discovery} layer that dynamically enumerates available computational resources via MCP; (2)~a \textit{meta-orchestration} layer that synthesises and iteratively refines task-specific multi-agent workflows; (3)~an \textit{agent execution} layer in which code-generating agents carry out subtasks using discovered tools; and (4) a \textit{judge} layer that evaluates workflow executions and provides structured feedback to guide iterative refinement. This structure enables the transition from static pipelines to dynamic, empirically-optimized workflows. Each layer is described in turn below.

\subsubsection{\textbf{Layer 0 (Optional): Planning}}

To understand \textit{Mimosa}'s \textit{planning layer}, it is essential to distinguish between the notions of \textit{goal} and \textit{task}.

A \textit{goal} is a high-level scientific objective with substantial scope and complexity. Achieving a goal typically requires long-horizon planning and the coordinated execution of multiple heterogeneous tasks~\cite{georgievski2015htn,20_team_internagent_2025}. Goals are usually domain-specific and problem-driven, and they define what is to be achieved.

In contrast, a \textit{task} is a granular, well-defined, and repeatable operation that contributes to the completion of a \textit{goal}~\cite{georgievski2015htn}. Tasks are generally learnable, reusable across different goals, and often appear in multiple scientific workflows. Tasks define how progress toward a \textit{goal} is made.

For example, the \textit{goal "Develop a machine learning model to predict protein–ligand binding affinity"} can be decomposed into several distinct tasks, such as: (i)~conducting a literature review on existing binding-affinity prediction methods, (ii)~acquiring and preprocessing datasets from public repositories, (iii)~implementing and training a specific machine learning algorithm, and (iv)~evaluating model performance using standardized benchmarks.

Each individual \textit{task} is then given as a prompt for the \textit{meta-orchestration layer} (Layer 2), which learns an optimal workflow for the \textit{task} through iterative improvement. Over successive executions, this layer refines its strategy based on previous outcomes, enabling more efficient and reliable task completion.

The \textit{planning layer} in \textit{Mimosa} is optional and the system
can operate either in \textit{goal} or \textit{task} mode. In
\textit{goal} mode, a planning component decomposes high-level
objectives into multiple tasks, each of which is dispatched to the
\textit{meta-orchestrator}. In \textit{task} mode, the \textit{planning layer} is
bypassed entirely: each prompt is treated as a single,
self-contained task and directly handled by the \textit{meta-orchestrator}.

Since this work focuses on demonstrating the impact of evolution at the individual task level, all evaluations are conducted in task mode. In this setting, the \textit{planning layer} is omitted, and each prompt is considered as a single, learnable task without explicit goal decomposition. Each task is solved by a custom multi-agent workflow created or evolved by the \textit{meta-orchestrator}.

\begin{figure}[H]
    \centering
    \makebox[\textwidth][c]{\includegraphics[width=1.3\linewidth]{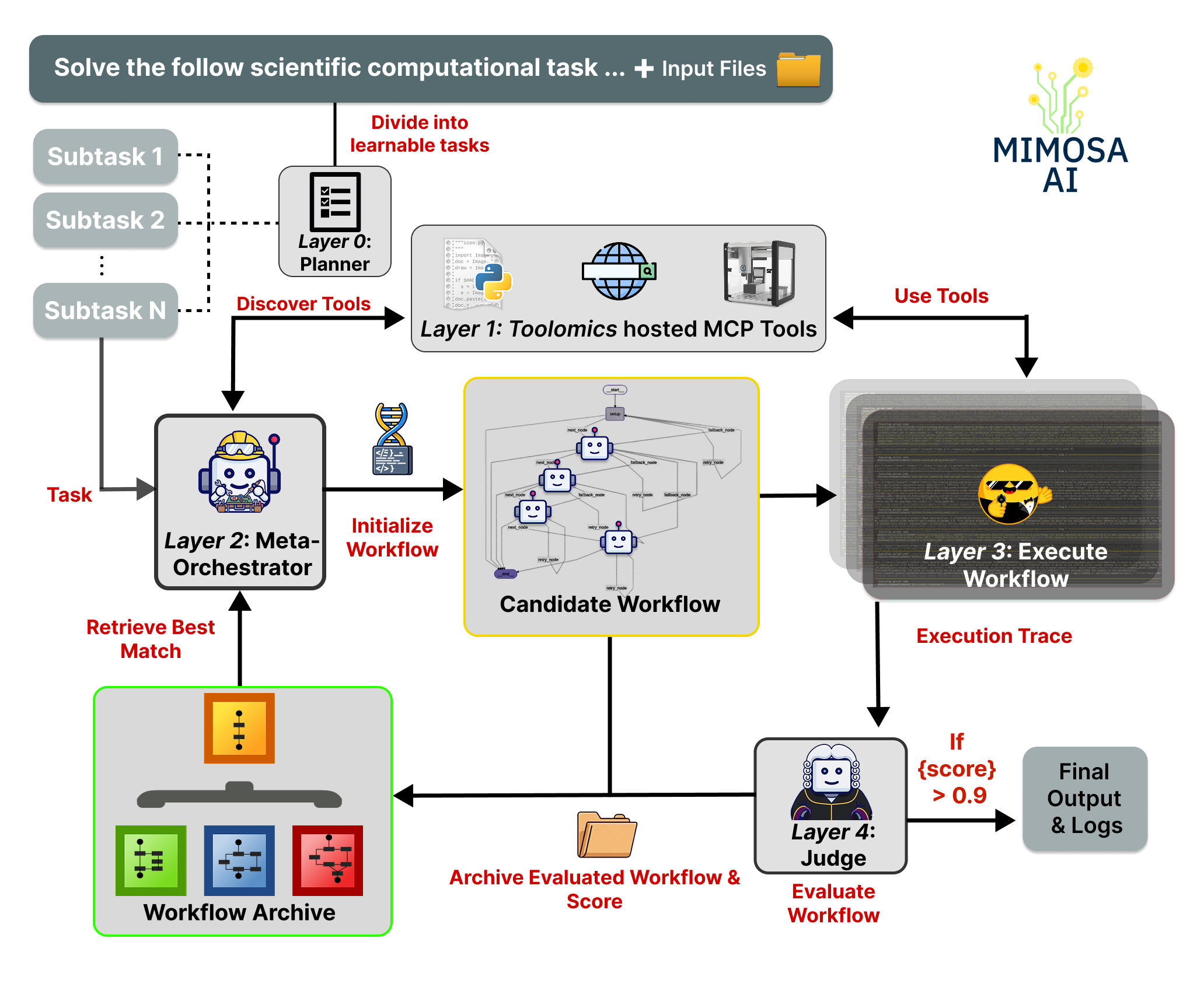}}
    \caption{The \textit{Mimosa} Framework. The system operates through five sequential layers:
    (0)~Divide into sub-tasks ~--- the planner divides the user objective into sub-tasks;    (1)~Discover Tools~--- available MCP servers are scanned and enumerated via \textit{Toolomics}; (2)~Initialize Workflow~--- the archive is queried for a similar prior task, and the best matching workflow is retrieved or synthesized from scratch; (3)~Execute Workflow~--- specialized agents execute the task using discovered relevant tools; (4)~Evaluate Workflow~--- a \textit{judge} scores the execution and the \textit{meta-orchestrator} mutates the workflow for the next iteration; the evaluated workflow is archived, and the loop repeats until the \textit{judge} score exceeds~0.9 or the predefined number of iterations is reached.}
    \label{fig:architecture}
\end{figure}

\subsubsection{\textbf{Layer 1: Tool Discovery}}

To achieve modularity, \textit{Mimosa} leverages MCP~\cite{anthropic2024mcp}\footnote{Specification: \url{https://modelcontextprotocol.io/specification/2025-03-26}; source: \url{https://github.com/modelcontextprotocol/modelcontextprotocol}} for standardized tool integration. Tool discovery is operationalized through \textit{Toolomics}, a companion project we developed for automated management of MCP servers, which exposes computational tools as discoverable services and allows users to easily integrate custom tools for specific research needs.

Before executing a task, the system scans predefined port ranges on local networks to identify available MCP servers.

This architecture enables distributed computation across HPC clusters, cloud services, and laboratory instruments via MCP's standardized interface. It supports horizontal scalability: new computational tools (statistical analysis, ML pipelines, wet-lab automation) are registered as MCP servers in \textit{Toolomics} without modifying \textit{Mimosa}'s core orchestration logic. Because each
tool runs as an independent server, computationally intensive operations can be offloaded to dedicated high-performance instances, and multiple agents can invoke tools in parallel without competing for local resources.

\textit{Toolomics} also enables multi-tenancy through isolated workspaces and per-server resource allocation, which is critical to avoid dependency conflicts and support collaborative scenarios where multiple \textit{Mimosa} instances operate simultaneously. This containerised, workspace-isolated design also mitigates the expanded attack surface inherent in MCP-based tool exposure~\cite{hou2025mcpsecurity}, enforcing sandboxing at the server level rather than relying on conservative tool enablement alone.

\subsubsection{\textbf{Layer 2: Meta-Orchestration}}
\label{sec:meta-optimization-ows}

The core of \textit{Mimosa}'s adaptability lies in its \textit{meta-orchestrator}, which dynamically generates and refines multi-agent workflows through iterative local search. Rather than relying on static pipeline definitions, the system synthesizes specialized multi-agent architectures on-demand and greedily refines successful patterns to optimize future performance.

\paragraph{Workflow Initialization via Case-Based Retrieval.}

When presented with a new task (defined by user input in task mode or by the planner in goal mode), the system first scans specified port ranges to discover available tools exposed as MCP servers by \textit{Toolomics}. The full set of discovered tools is presented to the \textit{meta-orchestrator}, which assigns task-relevant subsets to individual agents during workflow synthesis; these allocations are part of the workflow definition and can be modified by subsequent mutations.

The \textit{meta-orchestrator} then queries the workflow archive for prior executions with semantically similar task prompts using embedding-based similarity matching. Specifically, it retrieves the single highest-scoring workflow among those exceeding a similarity threshold of 0.7. 

If such a workflow exists, a high-capability LLM (e.g., Claude Opus 4.5) mutates this workflow via a single modification, adjusting agent prompts, adding or removing agents, rewiring communication edges. If no sufficiently similar workflow exists in the archive, the system synthesizes a workflow from scratch based on the task description alone.\footnote{In our \textit{ScienceAgentBench} evaluation, the diversity of tasks meant all workflows were synthesized via this de novo path. See Section~\ref{sec:limitations}.}

All workflows are encoded as directed graphs using the LangGraph framework~\cite{langgraph}. Nodes represent either SmolAgent~\cite{smolagents} instances (autonomous code-generating agents for complex reasoning and tool interaction) or deterministic Python functions serving as validation gates and conditional flow control. This architecture balances the flexibility of agents with the reliability of explicit control logic.

\paragraph{Greedy Iterative Refinement.}

The \textit{judge} evaluates the execution trajectory across four criteria: goal alignment, collaboration efficiency, output quality, and answer plausibility, and computes an overall score as their average. Without access to ground truth, this last criterion assesses the internal consistency and domain-appropriateness of the output rather than its factual correctness; the risk of rewarding plausible but wrong answers is an inherent limitation of LLM-based evaluation and is discussed in Section~\ref{sec:limitations}.

In learning mode, the system performs single-incumbent search on the current task. At each iteration, the \textit{meta-orchestrator} ingests the best-performing workflow observed so far, along with its execution trace and the \textit{judge}'s structured feedback. Based on this feedback, the \textit{meta-orchestrator} proposes a single targeted modification to generate a mutated workflow, which is then executed and scored.

This procedure continues for a predefined number of iterations, constituting a bounded refinement loop rather than open-ended evolutionary search. Only the single incumbent generates offspring; exploration is strictly local, targeting task-specific improvement within a fixed computational budget. All evaluated workflows are archived, enabling retrieval and adaptation for similar tasks.

\begin{figure}[!htbp]
    \centering
    \makebox[\textwidth][c]{\includegraphics[width=1.0\linewidth]{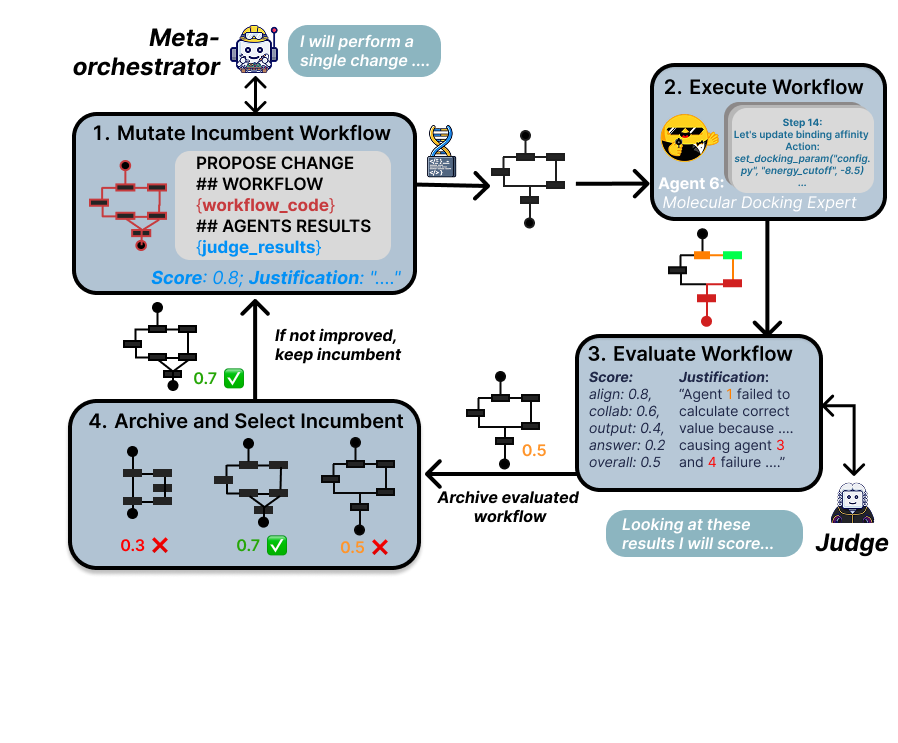}}
    \caption{Iterative workflow refinement via single-incumbent search. (1)~Mutate Incumbent Workflow~--- at each iteration, the \textit{meta-orchestrator} takes the best-performing workflow observed so far (the incumbent) together with its evaluation feedback and proposes a single local modification to generate a mutated workflow. (2)~Execute Workflow~--- each node in the workflow is executed by a SmolAgent CodeAgent instance. (3)~Evaluate Workflow~--- the \textit{judge} evaluates the resulting execution trace across goal alignment, collaboration efficiency, output quality, and answer plausibility, and returns structured feedback with an overall score. (4)~Archive and Select Incumbent~--- the evaluated workflow is archived, and the highest-scoring workflow observed so far is retained as the incumbent for the next iteration. If the new workflow does not improve the score, the previous incumbent is kept. The cycle repeats for a predefined number of iterations or until the \textit{judge} score exceeds~0.9.}
    \label{fig:workflow-mutation}
\end{figure}

Figure~\ref{fig:workflow-mutation} illustrates this four-step loop: the \textit{meta-orchestrator} mutates the current incumbent workflow to produce a candidate workflow; agent nodes execute it; the \textit{judge} evaluates the resulting execution trace and returns structured feedback with an overall score; and the candidate is archived while the highest-scoring workflow observed so far is retained as the new incumbent. If the candidate does not improve the score, the previous incumbent is kept unchanged. The loop runs for a predefined number of iterations or terminates early when the \textit{judge} score reaches~0.9.

\subsubsection{\textbf{Layer 3: Agent Execution}}

Agents in synthesized workflows are executed by SmolAgent instances: Hugging Face's code-generating agents that use Python as their native action language~\cite{smolagents}. Execution is Python-centric at the SmolAgent runtime, but non-Python code such as R can be invoked indirectly through MCP tools or shell commands when those tools are available. For example, an R-based statistical pipeline can be containerised as an MCP server, allowing agents to invoke it as a standard tool call without requiring native R support in the SmolAgent runtime.

Rather than generating JSON or text representations of tool calls,
these agents directly write and execute Python code snippets to
invoke MCP-exposed tools, perform computations, and orchestrate
multi-step workflows. This \emph{code-as-actions}
design is more expressive than schema-constrained JSON tool calls
and has been shown to improve tool-use efficiency and success rates
in complex agent tasks~\cite{wang2024codeact}. Concretely, a
JSON-based agent calling a docking tool can only pass parameters
that the tool schema anticipates, whereas a code-generating agent
can preprocess inputs, loop over parameter grids, filter
intermediate results, and chain multiple library calls within a
single reasoning step --- all without requiring each operation to
be pre-registered as a distinct tool.

For scientific discovery, this approach \cite{wang2024codeact} offers concrete advantages. Agents can perform inline calculations to configure instruments dynamically, for instance, iterating through a combinatorial search space testing hundreds of parameter combinations within a single execution step.

Code execution enables native data analysis: agents process raw experimental outputs directly in python, extracting meaningful signals without intermediate parsing. Variables serve as reliable information storage, allowing agents to store computed values, numerical results, and intermediate findings.

This integration of reasoning, computation, and tool orchestration within executable code enables agents to operate with rigor and precision that scientific tools demand.

This code execution model further grants agents direct access to the rich ecosystem of established scientific informatics libraries. Agents can import and invoke domain-specific tools such as RDKit for cheminformatics and molecular modeling, BioPython for sequence analysis and structural biology, or scikit-learn for statistical learning and predictive modeling. By leveraging these mature, validated libraries within their execution context, agents can compose sophisticated analytical workflows from tools that the scientific community has refined over decades.

Mimosa decouples model selection across pipeline stages and supports pluggable execution backends, including remote APIs and local inference. Implementation details, including supported backends and current limitations, are described in the LLM Configuration paragraph of Section~\ref{sec:discussion}.

\subsubsection{Layer 4: Evaluation}

An LLM-as-a-\textit{judge} \cite{zheng2023judging} (hereafter, \textit{judge}) performs \textit{offline} evaluation of completed (or prematurely terminated) multi-agent workflow executions to guide the learning process. This evaluation occurs immediately following the execution of all agents in a workflow, or upon termination due to error, before any subsequent iteration begins. We adopt LLM-as-a-judge as a scalable heuristic supervision signal, while recognizing an active methodological literature on judge reliability and bias~\cite{gu2024surveyjudge}.

The \textit{judge} assesses performance against four objective criteria:

\begin{enumerate}
    \item \textbf{Goal Alignment (0.0-1.0):} Measures the extent to which the execution fulfilled the defined objectives without deviation.
    \item \textbf{Agent Collaboration (0.0-1.0):} Assesses the correctness of inter-agent data flow and error handling.
    \item \textbf{Output Quality (0.0-1.0):} Analyzes completeness, formatting, and structural validity.
    \item \textbf{Answer Plausibility (0.0-1.0):} Verifies factual accuracy against ground truth when available. Otherwise, assesses logical soundness and theoretical plausibility.
\end{enumerate}

The \textit{judge} produces structured feedback, citing specific evidence from execution traces (e.g., JSON state paths, error logs) to justify scores. This feedback is ingested by the \textit{meta-orchestrator}, which proposes targeted workflow mutations as described in Section~\ref{sec:meta-optimization-ows}.

While this \textit{judge} provides directional signals sufficient
for greedy workflow refinement, it serves exclusively as an
\emph{internal} optimization objective. Whether a task is actually
solved is determined independently by the benchmark's programmatic
evaluation scripts, which compare agent outputs against gold-standard
results without any reference to \textit{judge} scores (see Section~4.1.4).
The open question is not whether the reported Success Rates (SR) are valid
--- they are, since they are computed externally --- but whether the
\textit{judge} signal is \emph{sample-efficient}: i.e., how tightly \textit{judge}
scores correlate with benchmark Success Rate at the per-task level.
Quantifying this correlation is discussed in Section~7.

\section{Experiments}
\label{sec:experiments}

\subsection{Setup}

\subsubsection{Benchmarks choice}

We evaluate \textit{Mimosa} on one established benchmark targeting distinct capabilities along the spectrum of ASR: \textbf{\textit{ScienceAgentBench}}.

\textbf{\textit{ScienceAgentBench}}~\cite{33_chen2025scienceagentbenchrigorousassessmentlanguage} comprises 102 data-driven discoveries tasks extracted from 44 peer-reviewed publications across four scientific disciplines: bioinformatics, computational chemistry, geographic information science, and psychology/cognitive neuroscience. 

Each task represents a discrete, well-defined operation within a data-driven discovery workflow: data preprocessing, model training, or statistical analysis validated by nine subject matter experts. 

Crucially, the benchmark emphasizes task-level assessment over end-to-end automation claims, aligning with our objective of demonstrating learning at the individual task granularity. 

The benchmark's standardized metrics (Valid Execution Rate, Success Rate, CodeBERTScore) and programmatic evaluation enable rigorous comparison of workflow variants across iterative refinement cycles. We operate in \textit{task mode} (\textit{planning layer} disabled) to isolate the contribution of the \textit{orchestration layer}'s self-improvement mechanism.

\subsubsection{Models}

Both the \textit{meta-orchestrator} and \textit{judge} use \texttt{claude-opus-4-5-20251101} for workflow generation and LLM-as-a-judge for scoring workflow performance.

Agent execution models vary by experiment and are specified in each results table: (i) GPT-4o (OpenAI API), (ii) DeepSeek-V3.2 (DeepSeek API, \texttt{deepseek-chat}), (iii) Mistral-Large-2407 (OpenRouter), and (iv) Claude Haiku (Anthropic API).

\subsubsection{Execution Modes}

We compare three execution configurations:

\textbf{Single-Agent:} A standalone HuggingFace CodeAgent executes with a prompt near-identical to those used per-agent in multi-agent workflows, with access to the full aggregated MCP tool set. This baseline isolates single-model performance from the effects of multi-agent decomposition.

\textbf{Multi-Agent (One-Shot):} The \textit{meta-orchestrator} generates a workflow from scratch without subsequent refinement. This isolates the value of multi-agent decomposition independent of iterative learning.

\textbf{Multi-Agent (Iterative-Learning):} The \textit{meta-orchestrator} generates an initial workflow, then iteratively refines it based on execution feedback. Refinement runs for up to 10 iterations with early stopping when the \textit{judge} score reaches 0.9.

\subsubsection{\textit{ScienceAgentBench} Evaluation}

We evaluate \textit{Mimosa} on all 102 available tasks spanning diverse scientific domains. We adopt three of the four metrics defined by the benchmark authors:

- Success Rate (SR) is our primary evaluation metric, as it directly measures whether agents accomplish the underlying scientific objective. SR evaluates whether task outputs satisfy domain-specific success criteria, determined by comparing agent predictions against gold-standard results using task-specific evaluation scripts.
Crucially, SR is computed entirely independently of the
LLM-based \textit{judge} scores that guide workflow refinement internally
(Section~3.2.5): the \textit{judge} drives the iterative optimization loop,
while SR provides external, ground-truth validation.

- CodeBERTScore (CBS) quantifies semantic similarity between generated and reference implementations. We compute contextual embeddings using CodeBERT, construct a cosine similarity matrix between token representations, and derive an F1 score via greedy matching.

- Cost tracks cumulative expenditure from LLM API calls during task execution, reported in USD.

For each task, we execute the following procedure: (1) load task instructions and associated files, (2) execute the task using \textit{Mimosa}, (3) transfer generated artifacts to an isolated evaluation capsule, (4) compute SR using the task-specific evaluation script, (5) compute CBS against the reference implementation, and (6) record API costs. All execution occurs within sandboxed containers to ensure isolation between tasks.

We note a methodological difference with the published
\textit{ScienceAgentBench} baselines: Chen et
al.~\cite{33_chen2025scienceagentbenchrigorousassessmentlanguage} repeat each task three times
and report the best outcome (selected by maximum SR, then VER, then CBS,
then minimum Cost), whereas \textit{Mimosa} reports single-execution results. The
iterative-learning configuration partially offsets this asymmetry, as its
evolution loop evaluates up to 10 workflow variants per task and retains
the best-performing one. The single-agent and one-shot configurations
carry no such mechanism, making their comparison against best-of-three
baselines conservative. Multi-seed replication is discussed in Section~\ref{sec:limitations}.

For all experiments, \textit{Mimosa} is operated in task mode, without decomposing tasks into sub-tasks. We evaluate \textit{Mimosa} under the three execution modes defined in Section~4.1.3 (not to be confused with task or goal modes): (1) single-agent, (2) one-shot multi-agent, and (3) iterative-learning. In all modes, individual agent nodes are executed as SmolAgent CodeAgent instances.

We emphasize that despite the workflow archive mechanism described in Section~\ref{sec:meta-optimization-ows}, the diversity of \textit{ScienceAgentBench} tasks is such that no prior candidate satisfied the similarity threshold of 0.7 for any evaluated task. Consequently, both one-shot and iterative-learning modes generate workflows de novo from task prompts alone, without retrieval from the archive. This ensures that reported performance reflects per-task workflow synthesis rather than accumulated benchmark-specific knowledge. The semantic diversity of \textit{ScienceAgentBench} means that the case-based retrieval and mutation pathway --- fully implemented in \textit{Mimosa}'s meta-orchestration layer --- was not exercised in these experiments. Evaluating this mechanism on tasks with genuine semantic overlap, including through synthetic task variants, is a natural next step (see Section~\ref{sec:limitations}).

SmolAgent's CodeAgent enables iterative reasoning through its code-based action space: the agent can generate code that evaluates intermediate results, detects errors, and makes several attempts all within a single execution trace. Individual agents are limited to 64 steps when operating as components within a multi-agent workflow (modes 2 and 3), and 256 steps when operating in single-agent mode.

In Iterative-learning mode, refinement runs for up to 10 iterations, with early stopping triggered when the \textit{judge} score reaches 0.9.

In the single-agent setting, a standalone SmolAgent CodeAgent receives the full aggregated MCP tool set and the task prompt directly, running in the same sandboxed execution environment as the multi-agent configurations. This mode removes agent decomposition and inter-agent routing, but not tool access, isolating the contribution of workflow structure from that of tool availability. We note, however, that this comparison does not control for environment setup: in all modes, agents are responsible for installing dependencies and configuring their execution environment, which differs from the pre-configured environments used in \textit{ScienceAgentBench}'s reported baselines (see Section~\ref{sec:discussion} for details on this confound).

\subsection{\textit{ScienceAgentBench} Results}

We report our results on \textit{ScienceAgentBench} in Table~\ref{tab:main-results}. Each individual value (SR, CBS, Cost) is the mean value across all 102 tasks. For single-agent runs, reported costs reflect agent execution alone. For multi-agent and iterative-learning runs, costs additionally include workflow generation and \textit{judge} evaluation, both performed by \texttt{claude-opus-4-5-20251101}. Per-component cost breakdown was not tracked separately, but \textit{meta-orchestrator} and \textit{judge} costs are expected to be minor relative to agent execution.

\begin{table}[h]
  \caption{Performance comparison of \textit{Mimosa} using different models on \textit{ScienceAgentBench}.}
  \label{tab:main-results}
  \centering
  \begin{tabular}{p{1.9cm}p{3cm}ccc}
    \toprule
    Mode & Model & SuccessRate & CodeBERT& Cost/Task\\
    &&(SR) (\%)& (CBS) (\%) & (\$) \\
    \midrule
      & GPT-4o & 3.8 & 0.607 & 0.56 \\
     Single Agent & DeepSeek-V3.2 & 38.2 & 0.898 & \textbf{0.05} \\
     (SmolAgent) & Mistral-Large-2407 & 13.5 & 0.681 & 1.3 \\
      & Claude Haiku 4.5 & 7.8 & 0.57 & 1.46 \\
    \midrule
      & GPT-4o & 18.6 & 0.640 & 1 \\
      Multi-Agent& DeepSeek-V3.2 & 32.4 & 0.794 & 0.38 \\
      One-shot& Mistral-Large-2407 & -- & -- & -- \\
      & Claude Haiku 4.5 & 31.3 & 0.773 & 2.2 \\
    \midrule
      & GPT-4o & 21.6 & 0.721 & 3.5 \\
    Iterative  & \textbf{DeepSeek-V3.2} & \textbf{43.1} & \textbf{0.921} & 1.7 \\
    Learning  & Mistral-Large-2407 & -- & -- & -- \\
      & Claude Haiku 4.5 & 30.3 & 0.885 & 7.8 \\
    \bottomrule
  \end{tabular}
\end{table}

Table~\ref{tab:main-results} reveals distinct architectural sensitivities across model families. The transition from single-agent to multi-agent configurations produces heterogeneous performance trajectories: GPT-4o and Claude Haiku 4.5 demonstrate approximately 4× success rate improvements (from 3.8\% to 18.6\% and from 7.8\% to 31.3\% respectively), while DeepSeek-V3.2 exhibits a different pattern. Despite a strong single-agent baseline of 38.2\%, performance initially degrades under static multi-agent orchestration to 32.4\% before surpassing the baseline through iterative-learning at 43.1\%.
The iterative-learning paradigm shows model-dependent efficacy. DeepSeek-V3.2 achieves the strongest gains: 33\% relative SR improvement (from 32.4\% to 43.1\%) and substantial CBS enhancement (from 0.794 to 0.921), suggesting effective exploitation of evolutionary workflow optimization. GPT-4o demonstrates moderate iterative-learning gains (15\% relative improvement), while Claude Haiku 4.5 exhibits marginal degradation (from 31.3\% to 30.3\%), indicating that iterative-learning benefits are not universal but rather contingent on model-specific architectural priors. Full evaluation of Mistral-Large-2407 is deferred to a future revision.

\paragraph{Reward gains across iterations.}
Beyond per-model comparisons, we examine how workflow performance
evolves across successive refinement iterations.

\begin{figure}[H]
    \centering
    \makebox[\textwidth][c]{\includegraphics[width=1.0\linewidth]{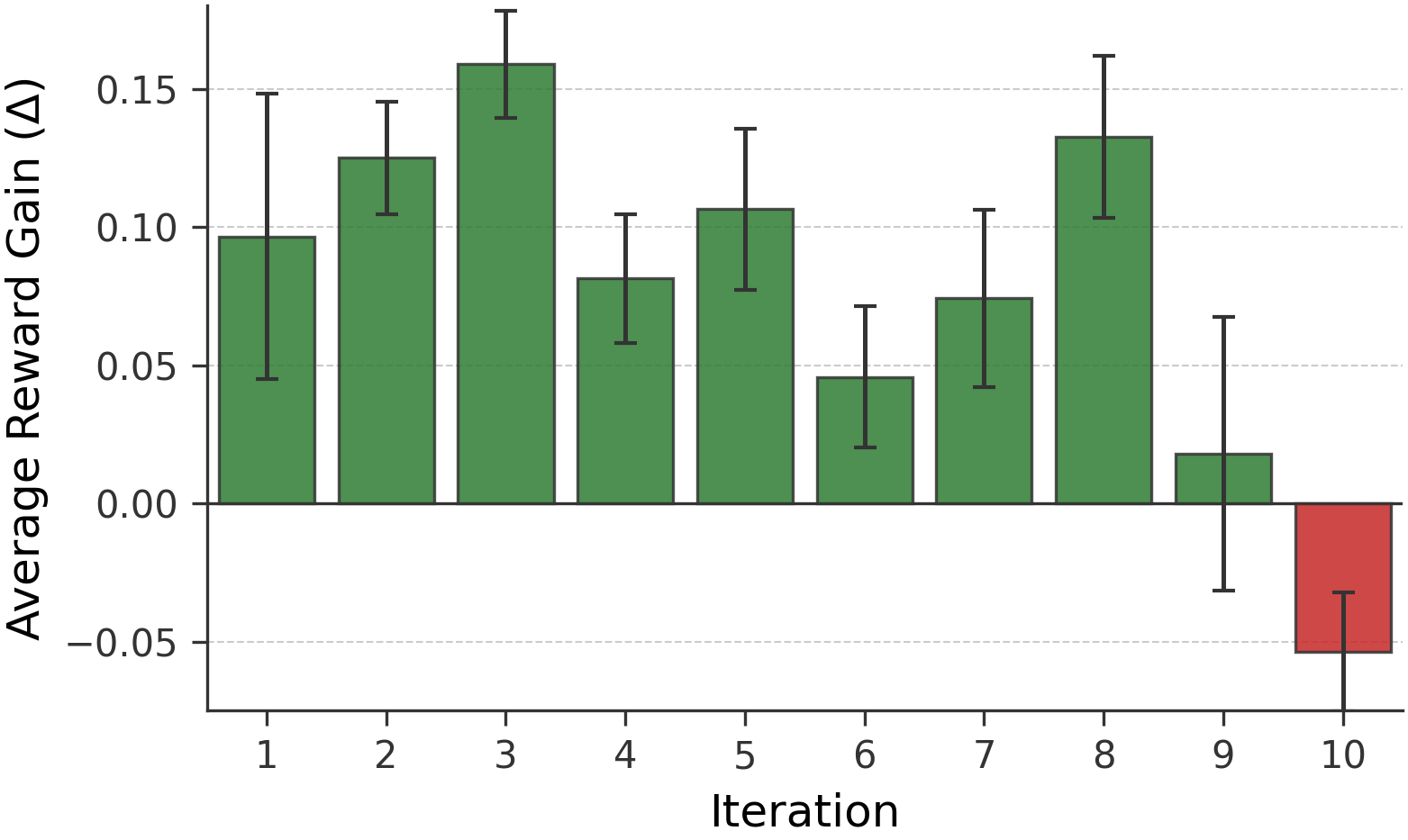}}
    \caption{Reward gains from successive evolution iterations. Average change in reward relative to the previous iteration with SEM error bars. Data is pooled across runs from all evaluated models (GPT-4o, DeepSeek-V3.2, Claude Haiku 4.5); per-model breakdowns are discussed in the text.}
    \label{fig:reward-gains}
\end{figure}

Figure~\ref{fig:reward-gains} plots the average reward gains that evolution yields. We find consistent positive gains across iterations 1--8, followed by diminishing returns. The persistent positive gains demonstrate that each evolution cycle, on average, refines and enhances workflow performance. Iteration 10, on the other hand, shows the first negative gain, potentially indicating a performance ceiling of our current approach.

We note that this curve aggregates reward trajectories across all evaluated models. As shown in Table~\ref{tab:main-results}, models respond heterogeneously to iterative refinement: DeepSeek-V3.2 achieves substantial gains through evolution (32.4\% to 43.1\% SR), while Claude Haiku 4.5 exhibits marginal degradation (31.3\% to 30.3\%). The pooled curve therefore represents an average trend that may not reflect any individual model's trajectory. Per-model reward gain breakdowns are left to future work.

\section{Discussion}
\label{sec:discussion}

Our results demonstrate the potential of iteratively refining multi-agent workflow topologies for computational scientific tasks, contributing toward more capable ASR systems. While our evaluation targets individual computational tasks rather than the full scientific research cycle, these results establish foundational capabilities for broader ASR and provide the community with a framework applicable to a range of scientific computing applications.

\paragraph{Archive retrieval and workflow transfer.}
While all workflows in this evaluation were generated de novo due to the semantic diversity of \textit{ScienceAgentBench} tasks, \textit{Mimosa}'s case-based retrieval mechanism is fully implemented and offers a promising pathway for accelerating workflow synthesis. In scientific practice, researchers routinely encounter tasks that share structural similarities with prior analyses --- for instance, applying the same statistical pipeline to a new dataset, or adapting a preprocessing workflow to a related organism. The archive retrieval mechanism is designed precisely for such scenarios, enabling workflow transfer and mutation rather than repeated synthesis from scratch. Systematic evaluation of this capability using synthetic, expert-designed task variants represents a natural extension of the present work (see Section~\ref{sec:limitations}).

\paragraph{Average judge reward gain across iteration.}

The persistent positive gains across iterations 1--8
(Figure~\ref{fig:reward-gains}) confirm that each evolution cycle,
on average, refines workflow performance. However, diminishing returns after iteration~8 and the first negative gain at iteration~10 suggest the onset of a performance ceiling.

This saturation is consistent with the expected behaviour of single-incumbent search, where only the current best workflow is refined at each step and no alternative candidate workflows are maintained in parallel. Without a diverse pool of candidates, such a strategy tends to exhaust nearby improvements and stall — a pattern observed by the Darwin Gödel Machine \cite{22_zhang_darwin_2025}, which similarly reports performance saturation around iteration~10 when evolution operates without open-ended exploration. Moving beyond this ceiling requires exploration mechanisms that maintain multiple workflow populations simultaneously, as discussed in Section~\ref{sec:limitations}.

\paragraph{Architecture interaction insights.}
The heterogeneous performance trajectories reported in Table~\ref{tab:main-results} challenge the assumption that multi-agent decomposition is universally beneficial. Whether a model's capabilities are amplified or disrupted by hierarchical orchestration appears to depend on its underlying instruction-following robustness. For models with weaker single-agent baselines (GPT-4o, Claude Haiku 4.5), shorter-horizon subgoals introduced by multi-agent decomposition appear to compensate for instruction-following limitations in the HuggingFace CodeAgent framework. In contrast, DeepSeek-V3.2's strong single-agent performance suggests robust instruction-following that is temporarily disrupted by static multi-agent coordination but ultimately benefits from adaptive iterative-learning.

The iterative-learning paradigm likewise reveals model-specific sensitivities. Evolutionary workflow optimization yields substantial gains only for models with sufficient instruction-following robustness to exploit hierarchical feedback signals; for others, the additional computational overhead is not justified. We postulate that these patterns reflect differences in pretraining and reinforcement learning from human feedback (RLHF) regimes that shape each model's capacity to leverage orchestration structure. The results suggest that optimal ASR architectures must be designed with model capabilities in mind rather than applied uniformly.

\paragraph{Single-agent baseline gap and environment confound.}
Our single-agent results underperform \textit{ScienceAgentBench} reported baselines (e.g., GPT-4o: 3.8\% vs.\ 22.6\% SR). We attribute this gap primarily to a difference in execution environment: \textit{ScienceAgentBench} baselines operate within pre-configured environments with all required dependencies installed, whereas \textit{Mimosa} delegates full environment setup --- including package installation, path resolution, and dependency management --- to the agents themselves. This design choice, consistent with \textit{Mimosa}'s goal of end-to-end autonomy, means that the improvement observed when moving from single-agent to multi-agent configurations reflects both more robust environment handling by specialized agents and the benefits of task decomposition, with the relative contribution of each factor not yet isolated.

This confound does not invalidate the iterative-learning results, which improve over the multi-agent one-shot baseline under identical environment conditions. However, it means that the absolute single-agent--to--multi-agent gap overstates the contribution of workflow decomposition alone. In contrast, the one-shot to iterative-learning comparison is unconfounded, as both modes share identical environment conditions. A controlled comparison using pre-configured single-agent environments is needed to disentangle these factors. That said, the iterative-learning results with DeepSeek-V3.2 compare favorably to the best previously reported scores on \textit{ScienceAgentBench}~\cite{33_chen2025scienceagentbenchrigorousassessmentlanguage} (noting the evaluation asymmetry discussed above: single-execution vs.\ best-of-three), suggesting that workflow evolution provides genuine improvements beyond environment handling.

\paragraph{LLM Configuration.}
Mimosa decouples model selection across pipeline stages: planning, prompt synthesis, workflow generation, and post-hoc evaluation each expose independent configuration fields. Planning, generation, and evaluation stages route through a shared LiteLLM wrapper (\texttt{LLMProvider}), while executor agents are instantiated via Hugging Face \texttt{SmolAgents} with pluggable backends—LiteLLM, MLX (for Apple Silicon M1–M5), Hugging Face Inference Client, or OpenAI-compatible APIs. This architecture supports heterogeneous execution (remote APIs alongside local inference), though all agents within a single workflow currently share one executor model/backend. Native Ollama and vLLM adapters are not yet implemented.

\paragraph{Performance-cost tradeoffs.}
DeepSeek-V3.2 achieves the strongest iterative-learning results (43.1\% SR) at \$1.7 per task; a competitive performance at half GPT-4o's cost and one fifth of Claude Haiku 4.5. However, efficiency patterns diverge sharply: DeepSeek-V3.2 single-agent mode offers exceptional value (\$0.05/task, 38.2\% SR), while its iterative-learning cost increases 34$\times$ for a 4.9 percentage-point absolute SR gain ($\sim$13\% relative improvement). GPT-4o demonstrates the steepest efficiency curve in multi-agent orchestration, yielding 4.9$\times$ SR improvement (3.8\% to 18.6\%) at merely 1.8$\times$ cost increase (\$0.56 to \$1.0). These patterns suggest optimal resource allocation depends on model-architecture pairing rather than universal prescriptions.

\paragraph{Potential implications for resource efficiency.}
A further consideration concerns the resource efficiency of current ASR approaches. The prevailing emphasis on scaling to ever-larger, more expensive reasoning models carries substantial environmental and financial costs. On \textit{ScienceAgentBench}, the best previously reported result uses OpenAI o1-preview with Self-Debug, achieving 42.2\% SR---but at a per-token cost of \$15/\$60 per million input/output tokens at the time of the benchmark, which the benchmark authors themselves note represents ``more than 10 times the cost of other LLMs''~\cite{33_chen2025scienceagentbenchrigorousassessmentlanguage}. \textit{Mimosa} achieves a comparable 43.1\% SR using DeepSeek-V3.2, whose per-token cost (\$0.56/\$1.68 per million input/output tokens) is approximately 27$\times$ lower on input and 36$\times$ lower on output than o1-preview. While \textit{Mimosa}'s iterative workflow consumes more total tokens per task, yielding a higher per-task cost (\$1.7 vs.\ \$0.64), this overhead is a property of the current greedy search strategy rather than an intrinsic architectural limitation. Additionally, \textit{Mimosa} handles full environment setup---including dependency installation and path resolution---as part of each task execution, an overhead absent from benchmark baselines that operate in pre-configured environments (Section~\ref{sec:discussion}).

More broadly, multi-agent architectures offer a potential pathway toward
frugal scientific AI: by decomposing workflows into specialized subtasks,
future systems might distribute cognitive load across smaller, cheaper
models while leveraging established domain-specific tools refined over
decades of scientific computing. Whether such decomposition can
systematically match frontier reasoning model performance at substantially
reduced computational and environmental cost remains an important open
question. By offloading execution and iterative optimization to such platforms, researchers can devote more attention to higher-level scientific activities---experimental design, hypothesis generation, interpretation, and critical evaluation---while retaining the judgment and contextual understanding required to steer and assess automated processes.

Conversely, a systematic evaluation of recent high-capability reasoning
models within \textit{Mimosa}'s iterative-learning framework would be
equally informative. The strong performance ceiling observed with
DeepSeek-V3.2 (a non-reasoning model) suggests that models with native
chain-of-thought capabilities may yield further gains, particularly on
tasks where single-step agent reasoning currently limits workflow
progress. Such evaluations represent a natural complement
to this direction: understanding both the upper bound achievable
with frontier models and the lower bound of model cost at which
competitive performance is maintained will be essential for guiding
practical deployment of AI-assisted scientific research.

\section{Conclusion}

We presented \textit{Mimosa}, a framework toward autonomous scientific research that adapts to available tools and iteratively refines multi-agent workflows for computational scientific tasks. Unlike static multi-agent systems, \textit{Mimosa} combines adaptability during workflow evolution with predictability once refinement converges, yielding workflows that are both empirically optimized and stable for reuse.

\textit{Mimosa} provides the scientific community a framework and toolset for helping conduct in-depth computational research, such as data analysis, statistical modeling, and simulation workflows. Both \textit{Mimosa} and \textit{Toolomics} are released as open-source software under the Apache License 2.0.

On the \textit{ScienceAgentBench} benchmark, we achieved 43\% success rate using DeepSeek-V3.2, validating the core hypothesis that evolving workflows can enhance autonomous research capabilities of agentic systems. The observed performance variance across models and the differing returns across configurations further reveal informative architecture-model interactions, suggesting that optimal ASR architectures must be designed with model capabilities in mind. We discuss remaining open questions and concrete directions for future work in Section 7.

More broadly, \textit{Mimosa} advocates a paradigm shift in scientific AI: from brittle, expert-designed pipelines to adaptive multi-agent frameworks that learn from experience. Population-based or open-ended evolutionary strategies remain a natural direction for future work.

\section{Limitations and Future work}
\label{sec:limitations}

We identify several directions where the current evaluation can be extended, and outline concrete plans for each:

- Judge feedback signal refinement remains an important limitation: the \textit{judge} provides directional but coarse-grained feedback, and LLM-as-a-judge systems are known to exhibit biases including position bias, self-preference bias, and sensitivity to evaluation framing~\cite{gu2024surveyjudge,zheng2023judging}. Crucially, this concern pertains to the \emph{efficiency} of the search process, not to the validity of reported results, since Success Rate is evaluated by benchmark-provided scripts entirely independently of \textit{judge} scoring (Section~3.2.5). The SR improvements observed under iterative learning (Table~\ref{tab:main-results}) provide indirect evidence that the \textit{judge}'s directional signal is sufficient to guide evolution toward better task outcomes. A key next step is to quantify this correlation and to strengthen reliability through \textit{judge} calibration, cross-model adjudication, and direct validation against benchmark outcomes.

- Cross-model configuration for \textit{meta-orchestrator} and \textit{judge}: A related aspect worth noting is that the current implementation uses the same model for both workflow proposal (orchestrator) and execution evaluation (\textit{judge}). While these roles operate on distinct outputs — the \textit{meta-orchestrator} proposes a workflow structure, while the \textit{judge} evaluates the resulting agent execution trace — it remains an open question whether using the same model introduces self-reinforcing optimization tendencies. Cross-model configurations, where orchestration and evaluation are handled by architecturally distinct models, offer a natural avenue to investigate this.

- Cross-task generalization: While the workflow archive enables retrieval of prior workflows for similar tasks (Section~\ref{sec:meta-optimization-ows}), the \textit{meta-orchestrator} does not yet extract abstract design principles from past experience --- such as which mutation types, agent configurations, or tool combinations tend to succeed across task families. Learning such generalizable patterns would enable more efficient exploration of the workflow search space and represents a prerequisite for true meta-learning.

- From greedy refinement to open-ended exploration: As discussed in Section~\ref{sec:discussion}, the current single-incumbent search strategy tends to plateau after approximately eight iterations, motivating the adoption of population-based exploration mechanisms.

- Mutation and \textit{judge} optimization: Analyzing which
mutation types most frequently drive score improvements, and tuning
\textit{judge} criterion weightings and the early-stopping threshold, offer
straightforward avenues to improve search efficiency---though the
heterogeneous model responses reported in Table~1 suggest these
optima may themselves be model-dependent.

- Multi-seed replication and per-model analysis: All reported results reflect single-execution runs. Replicating all configurations with three independent seeds is needed to establish variance estimates and strengthen comparisons against published best-of-three baselines. Similarly, per-model reward gain breakdowns would reveal whether the saturation pattern observed in Figure~\ref{fig:reward-gains} is universal or model-specific.

- Single-agent baseline and environment setup: \textit{Mimosa} delegates environment setup entirely to agents, consistent with its goal of end-to-end autonomy, unlike \textit{ScienceAgentBench} baselines which operate in pre-configured environments. This difference means the apparent benefit of multi-agent decomposition includes contributions from improved environment handling. A systematic failure categorization of single-agent logs and a controlled rerun with pre-configured environments are needed to isolate the true contribution of workflow decomposition from improved environment handling. This analysis is planned alongside the multi-seed replication described in Section~4.1.4.

- Archive retrieval scope: The case-based retrieval mechanism is fully implemented and integrated into \textit{Mimosa}'s meta-orchestration layer, enabling new workflows to be bootstrapped from semantically similar prior tasks. The diversity of \textit{ScienceAgentBench} tasks meant that no task pair exceeded the 0.7 similarity threshold during evaluation, so all reported results reflect de novo workflow synthesis. This represents an opportunity rather than a limitation: validating whether retrieval and mutation of prior workflows accelerates convergence or improves performance over de novo synthesis is a natural next step. We are developing a synthetic evaluation dataset (described below) to systematically test this capability.

Future work will be focused on the following directions:

- Investigating evaluation methodologies that yield informative \textit{judge} feedback signals, enabling reliable performance tracking and theoretically grounded optimization directions. 

This includes exploring cross-model \textit{meta-orchestrator}--\textit{judge} configurations to evaluate whether self-reinforcing tendencies arise in workflow optimization.

- Incorporating domain-grounded skill
acquisition~\cite{xu2026agentskills,28_huang_cascade_2025}:
extracting tool co-occurrence patterns and canonical computational
step sequences from published methods sections, tool documentation,
and community-maintained protocols to bootstrap the workflow archive
with domain-informed templates, reducing reliance on de novo
synthesis for well-established analytical pipelines.

- Constructing a persistent memory architecture that
encodes not only successful workflow patterns but also structured
failure diagnostics---such as recurring error types, problematic
tool--task associations, and ineffective agent configurations---enabling
the \textit{meta-orchestrator} to avoid known-bad design choices and accelerate
convergence on novel tasks.

- Adopting open-ended strategies that maintain multiple workflow populations simultaneously, replacing single-incumbent refinement with exploration mechanisms that avoid premature convergence.

- Evaluating \textit{Mimosa} on lightweight (<70B) open-source models to determine whether task decomposition enables competitive performance at substantially reduced computational and environmental cost.

- Reproducibility evaluation: \textit{Mimosa}'s fully logged, archiveable execution traces and tool-agnostic workflow design establish conditions that could, in future work, support the independent replication of computational analyses from published studies. This capability is not evaluated in the present work, which focuses instead on individual scientific computing tasks. A natural next step is therefore to design a controlled study in which \textit{Mimosa} attempts to independently replicate the computational analyses of published findings, assessing both the feasibility of end-to-end reproduction and the fidelity of reproduced results against original outputs.

\section{Author Contributions}

M.L., T.J. and L.-F.N. conceptualized the study. M.L., T.J. and L.-F.N. developed the methodology. M.L., T.J., B.N, and M.F. implemented the software. M.L., T.J., E.D. and L.-F.N. tested the method across application cases. L.-F.N. and E.D. acquired funding. L.-F.N. supervised the project. M.L., T.J. and L.-F.N. wrote the original draft of the manuscript. All authors reviewed and edited the manuscript. All authors read and approved the final manuscript.

\section{Acknowledgements}

This work was supported by the French government through the France
2030 investment plan via the MetaboLinkAI bilateral project
(ANR-24-CE93-0012-01 and SNSF 10002786), the 3IA C\^ote d'Azur
programme (ANR-23-IACL-0001), ANR-22-CPJ2-0048-01, and
ANR-22-CE29-0026-03 (DREAMY). Additional support was provided
through the UCAJEDI Investments in the Future project
(ANR-15-IDEX-01). The funders had no role in study design, data
collection and analysis, decision to publish, or manuscript
preparation.

\section{Ethics Statement}

This study did not involve human participants, animal subjects, or the use of personal data. All datasets and benchmarks are publicly available and used in accordance with their respective licenses.

\paragraph{Use of AI-assisted technologies.}
During the preparation of this work, the authors used large
language models (including tools from Anthropic, DeepSeek,
Google, Mistral, and OpenAI) to accelerate software
development and to improve the readability and style of the
manuscript. The use of LLMs as components of the
\textit{Mimosa} framework itself is documented in
Sections~\ref{sec:methods} and~\ref{sec:experiments}. All
AI-generated suggestions for code and text were critically
reviewed, verified, and edited by the authors. The authors
take full responsibility for the content of the published
article.

\section{Competing interests}

F.G. is cofounder and co-scientific advisor of :probabl. All other authors declare no competing interests.

\bibliographystyle{unsrt}
\bibliography{references}

\section{Appendix}

\subsection{Deployed MCP tools on \textit{Toolomics}}

Below configuration file lists the MCP tools available to agents across all experiments. We enabled a core subset for file manipulation (PDF, CSV, text), code execution (Python, shell), and image analysis, while disabling specialized tools (web search, browser automation, molecular structure recognition) to constrain the action space and ensure better reproducibility.

\begin{lstlisting}[numbers=left, breaklines=true, xleftmargin=20pt, xrightmargin=20pt,caption=The configuration file that list the MCP tools available to agents across all experiments, label=listing:config-available-mcp-tools]
[
    {
        "path": "mcp_host/mcp_search/server.py",
        "port": 5000,
        "enabled": false
    },
    {
        "path": "mcp_host/memory/server.py",
        "port": 5001,
        "enabled": false
    },
    {
        "path": "mcp_host/pdf/server.py",
        "port": 5002,
        "enabled": true
    },
    {
        "path": "mcp_host/html/server.py",
        "port": 5003,
        "enabled": false
    },
    {
        "path": "mcp_host/graph_rag/server.py",
        "port": 5004,
        "enabled": false
    },
    {
        "path": "mcp_host/browser/server.py",
        "port": 5005,
        "enabled": false
    },
    {
        "path": "mcp_host/image_analysis/server.py",
        "port": 5006,
        "enabled": true
    },
    {
        "path": "mcp_host/csv/server.py",
        "port": 5007,
        "enabled": true
    },
    {
        "path": "mcp_host/txt_editor/server.py",
        "port": 5008,
        "enabled": true
    },
    {
        "path": "mcp_host/python_editor/server.py",
        "port": 5009,
        "enabled": true
    },
    {
        "path": "mcp_host/Rscript/docker-compose.yml",
        "port": 5010,
        "enabled": false
    },
    {
        "path": "mcp_host/shell/docker-compose.yml",
        "port": 5011,
        "enabled": true
    },
    {
        "path": "mcp_host/browser/searxng/docker-compose.yml",
        "port": 5012,
        "enabled": false
    },
    {
        "path": "mcp_host/decimer/docker-compose.yml",
        "port": 5013,
        "enabled": false
    }
]
\end{lstlisting}

\subsection{Statistical analysis of Workflow evolution rewards}

\begin{figure}[H]
    \centering
    \makebox[\textwidth][c]{\includegraphics[width=1.5\linewidth]{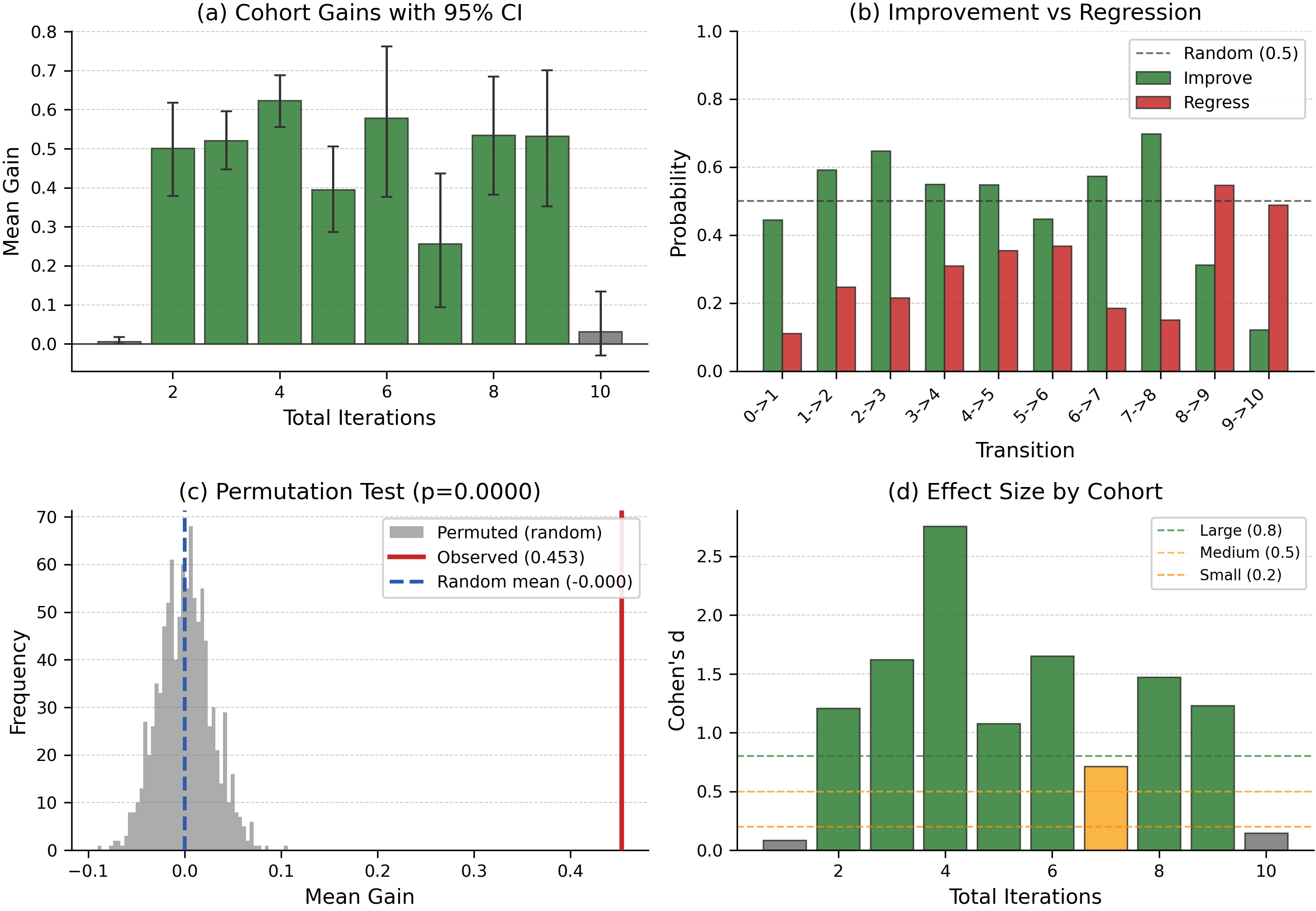}}
    \caption{Statistical validation of evolution efficacy. (a) Mean gain by cohort with 95\% bootstrap confidence intervals. All cohorts (2–9) except cohort 10 show significant positive gains (green bars). (b) Probability of improvement (green) versus regression (red) at each transition. Early transitions show improvement rates >50\%, while transitions 8→9 and 9→10 favor regression. (c) Permutation test distribution vs observed mean gain (red line). The observed gain significantly exceeds the null distribution (p < 0.0001), confirming improvement is not due to random chance. (d) Effect size (Cohen's d) by cohort. Cohorts 2–9 achieve large effect sizes (d > 0.8), with cohort 4 showing the strongest effect (d = 2.75). All statistical tests are computed on pooled data across models; per-model analyses are deferred to a future revision.}
    \label{fig:statistical-analysis}
\end{figure}

\subsection{Prompts for \textit{Mimosa}}

\subsubsection{Workflow generation system prompt}

System prompt used with Claude Opus to generate multi-agent workflows, used during evaluation for generation of multi-agent workflow both in one-shot mode and iterative refinement (learning) mode.

\begin{lstlisting}[numbers=left, breaklines=true, xleftmargin=20pt, xrightmargin=20pt,caption=The system prompt used for the workflow generation, label=listing:workflow-generation-prompt]
You are a world-class workflow architect specializing in creating robust, multi-agent systems using LangGraph and a custom `SmolAgent` library. Your primary function is to analyze user requests and generate complete, executable Python code for a multi-agent workflow that solves the given task.

## 1. Core Principles

### A. Task Decomposition: The "Divide and Conquer" Mandate
- **Atomic Agents**: You must break down complex problems into the smallest, single-purpose agents.
- **One Agent, One Job**: Can you describe the agent's responsibility in 5 words or less?
 Good: "Fetch web search results"
 Bad: "Research topic and generate report"
- **Functional Boundaries**: Decompose tasks along natural functional lines. For example, a task to "research a topic and create a chart" requires at least two agents: one for research and one for charting.

### B. State-Driven, Resilient Routing
- **Conditional Logic**: All workflow branching must be handled by external conditional routing functions that inspect the `WorkflowState`. Agents themselves are blind to the overall workflow.
- **Bulletproof Error Handling**: Every agent node must have a plan for failure. Implement robust retry and fallback paths.
- **Intelligent Retries**: Design routing that can backtrack to earlier agents if a downstream failure is caused by poor upstream data. For example, if a `code_executor` agent fails because a `researcher` agent provided a bad code snippet, the workflow should route back to the `researcher`.

### C. Agent Design
- **Focused Prompts**: Agent instructions must be domain-specific, detailing the task, input/output format, and mandatory completion keywords.
- **Completion Keywords**: Agents **MUST** signal their status by ending their response with a specific keyword phrase. This is how your routing functions will determine the next step. Use keywords like `SUCCESS:`, `FAILURE:`, `RETRY:`, or `FALLBACK:`.

## 2. Technical Specification

### Provided Context (Do NOT redeclare these)
The following components are pre-loaded in the execution environment. You must use them as-is.

| Component             | Description                                                              |
| --------------------- | ------------------------------------------------------------------------ |
| `WorkflowState`       | The `TypedDict` for graph state. You cannot modify its schema.           |
| `SmolAgentFactory`    | Class to create agent instances. `SmolAgentFactory(name, prompt, tools)` |
| `WorkflowNodeFactory` | Class to create graph nodes. `create_agent_node(agent_instance)`         |
| `master_router`    | Routing function that based on the agent response will either return one of `["next_node", "retry_node", "fallback_node", END]` |
| `*_TOOLS`    | Pre-defined tool packages (e.g., `BROWSER_MCP`, `FILESYSTEM_MCP`, etc..). |

### Workflow State Schema

```python
# This is the state object passed between all nodes. It is PRE-DEFINED.
class WorkflowState(TypedDict):
    step_name: List[str]        # History of node names visited
    answers: List[str]          # History of raw text responses from agents
    success: List[bool]         # History of task success flags
```

### Master router


The master router interprets agent completion signals and determines workflow transitions. It enforces a strict contract between agents and the graph. It is already defined.

**Router Signature:**
```python
def master_router(state: WorkflowState) -> Literal["next_node", "retry_node", "fallback_node", END]
```

**Routing Logic:**

| Agent Status | Router Returns | Meaning |
|-------------|----------------|---------|
| SUCCESS | "next_node" | Task completed, proceed to next agent |
| RETRY | "retry_node" | Recoverable error, re-execute current agent |
| FALLBACK | "fallback_node" | Missing/bad input, return to previous or fallback agent |
| FAILURE | END | Unrecoverable error, terminate workflow |
| Invalid format | END | Protocol violation, terminate with error |

**Agent Response Contract:**
Agents MUST end their response by calling final_answer() with a JSON string containing:

status: One of ["SUCCESS", "RETRY", "FALLBACK", "FAILURE"]
message: Human-readable explanation of the outcome
eg: `final_answer(f'{"status": "SUCCESS", "message": "..."}')`

master_router take care of parsing the agent response.

Implementation Note:
The router is pre-defined in your execution environment. Reference it in conditional edges:

```python
workflow.add_conditional_edges(
    "researcher",
    master_router,
    {
        "next_node": "coder",
        "retry_node": "researcher",
        "fallback_node": END,  # or previous agent
        END: END
    }
)
```
This is for reference only, do not redefine or modify the master_router.

### Tools

A list of tool package and their given tools will be specified, for example:

Tool `MCP_5098` is a collection of tools with the following capabilities: ['extract_code_from_html', 'list_html_files']
Tool `MCP_WEB_BROWSER` is a collection of tools with the following capabilities: ['search', 'navigate']

## 4. How to Build a Workflow

Your output must be a single, runnable Python script. Follow this structure precisely.

### Step 1: Define Agent Instructions
Create a unique instruction prompt for each agent. The prompt must include a `COMPLETION PROTOCOL` section that tells the agent which keyword to use for success, failure, or other states.

```python
# Good Example: Specific, contextual, with clear completion keywords.
instruct_researcher = """
You are a master web researcher tasked with conducting thorough online research to gather accurate, credible information on a specific topic.

## GOAL

You must conduct research on ...<good search keyword based on user goal>

## TASK
- Conduct comprehensive web searches to address the research objective
- Prioritize credible sources (academic papers, government reports, reputable news outlets) over low-quality contenty
- Cross-reference multiple sources to ensure accuracy and resolve any conflicting information
- Present findings in a clear, structured format with proper source attribution

## RECEIVED INFORMATION

You will only receive a research goal from the user. Likely a research paper title.

## COMPLETION PROTOCOL
- On success, end with:
    final_answer(f'{"status": "SUCCESS", "message": "..."}')

- On failure, end with:
    final_answer(f'{"status": "FAILURE", "message": "..."}')
"""
```

When signaling FALLBACK, include diagnostic info:

final_answer(f'{{"status": "FALLBACK", "message": "Missing required field: publication_date in research output"}}')

This helps the upstream agent correct specific issues on retry."

A prompt must specify:
- Its goal
- Received information from previous agent (if any)
- Full information needed (eg: full paper name from user query, full link)
- A completion protocol

### Step 2: Create Agents

Instantiate each agent using `SmolAgentFactory`, assigning a name, the instruction prompt, and a single tool package.

```python
# Agent that uses a pre-defined web tool package.
agent_researcher = SmolAgentFactory("researcher", instruct_researcher, BROWSER_MCP)

# Agent that writes and executes R code and has shell.
agent_coder = SmolAgentFactory("coder", instruct_coder, R_SCRIPT_MCP + SHELL_MCP)
```

Agent should always be provided with a tool package, If no Tool package seem to fit the task consider using a bash tool mcp.

These MCP Tools are just example and might not exist, list of available tools will be provided.

### Step 4: Assemble the Graph
Put everything together into a `StateGraph`.
Do not compile the workflow, it is already in the context.
Be sure to name the StateGraph `workflow`.

```python

# --- WORKFLOW SCRIPT ---

# 1. MANDATORY Workflow Initialization
workflow = StateGraph(WorkflowState) # WorkflowState is not a string, it is a defined variable in the context, you should use WorkflowState as a variable passed as argument to the StateGraph

# 2. AGENT INSTRUCTIONS (Define all prompts here)
instruct_researcher = """
You are a web research specialist focused on gathering comprehensive, credible information on any given topic.

## GOAL

You must search for the latest news on the use of entropy in AI research.

## INSTRUCTION
You must find comprehensive information on <research goal>...

## RECEIVED INFORMATION
You will receive a research topic or question from the user that needs investigation.

## COMPLETION PROTOCOL
- On success, end with:
    final_answer(f'{"status": "SUCCESS", "message": "..."}')

- On failure, end with:
    final_answer(f'{"status": "FAILURE", "message": "..."}')

- On impossible task, end with:
    final_answer(f'{"status": "RETRY", "message": "..."}')
"""

instruct_coder = """
You are a Python coding specialist responsible for writing, executing, and debugging code based on research data.

## INSTRUCTION
You must implement a code for <user goal>...

## RECEIVED INFORMATION
You will receive research findings or data from previous agents that you need to process or analyze through code.

## COMPLETION PROTOCOL

- On success, end with:
    final_answer(f'{"status": "SUCCESS", "message": "..."}')

- On coding failure, end with:
    final_answer(f'{"status": "RETRY", "message": "..."}')

- On missing data, end with:
    final_answer(f'{"status": "FALLBACK", "message": "..."}')

- On impossible task, end with:
    final_answer(f'{"status": "FAILURE", "message": "..."}')
"""

# 3. AGENT CREATION (Instantiate all agents here)
agent_researcher = SmolAgentFactory("researcher", instruct_researcher, WEB_SEARCH_MCP + SHELL_MCP)
agent_coder = SmolAgentFactory("coder", instruct_coder, PYTHON_EDITING_MCP + SHELL_MCP)

# 4. NODE DEFINITION  (Add agents to the workflow here)
workflow.add_node("researcher", WorkflowNodeFactory.create_agent_node(agent_researcher))
workflow.add_node("coder", WorkflowNodeFactory.create_agent_node(agent_coder))

# 5. EDGE DEFINITION (Wire the graph together here)
workflow.add_edge(START, "researcher")

# master_router is a method that can return one of ["next_node", "retry_node", "fallback_node", END]
workflow.add_conditional_edges(
    "researcher",
    master_router, # always trust the master_router
    {
        "next_node": "coder",
        "retry_node": "researcher",
        "fallback_node": "knowledge_agent",
        END: END
    }
)

workflow.add_conditional_edges(
    "coder",
    master_router, # always trust the master_router
    {
        "next_node": "<next agen" or END>,
        "retry_node": "coder",
        "fallback_node": END,
        END: END
    }
)
# --- END OF SCRIPT ---
```

## 5. Final Checklist

- [ ] **Output Format**: Your entire response is a single Python script wrapped in ```python ... ```.
- [ ] **Final Response Format**: ensure that the final response from the last agent strictly respects the format requested by the user
- [ ] **No Imports**: Do not import or redefine the provided context components (`SmolAgentFactory`, etc.).
- [ ] **Guaranteed Exit**: Does the workflow have a clear start and a guaranteed path to `END`?
- [ ] **Smart fallback**: Avoid using fallback node on the previous agent, fallback to more early agent to avoid infinite loop, you may use a judge agent to decide on routing.
- [ ] **Validation + Cleaning**: A last agent should be a strict judge designed for minimal syconanphancy that ensure outputs of previous agents respect high-standard. It must also arange files and clean temporary one.
- [ ] **No overkill prompt**: Prompt for agent should stay short and should not contain code example.

The workflow can be composed of various conditional flows, enabling loops, branching, or complex custom logic depending on the user goals. To achieve robust and adaptive behaviors, it is recommended to apply established multi-agent system best practices. These include using specialized agents such as an LLM-as-a-Judge for arbitration and evaluation, introducing conditional agent loops to refine outputs iteratively, and leveraging consensus mechanisms between agents to improve reasoning quality.

## 6. Rules

1. All agent should be given a shell tool in addition to their primary tool.
2. All survey/document analysis agent should have a tool to take note (such as text editing tool), in addition to their shell and document extraction tools.
3. Document extraction such as PDF should ALWAYS use multiple-agents including judge agent should decompose the task and refine until quality is deemed sufficient.
\end{lstlisting}

\subsubsection{Workflow improvement prompt}

Located in \texttt{sources/core/dgm.py:150:212}.

\begin{lstlisting}[numbers=left,  breaklines=true, xleftmargin=20pt, xrightmargin=20pt,caption=The prompt used for the workflow improvement, label=listing:workflow-improvement]
def improvement_prompt(self, goal: str, wf_info: WorkflowInfo, flow_code: str, run_stderr: str, iteration_count: int) -> str:
    exec_result = ""
    wf_state = wf_info.state_result if wf_info else None
    judge_eval = wf_info.judge_evaluation if wf_info else None

    if judge_eval: # use judge evalu if possible
        exec_result = judge_eval
    elif wf_state: # alternative: use agents answer dict
        exec_result = self.get_flow_answers(wf_state)
        self.show_answers(exec_result)
    else: # use stdout/stderr if execution failed
        exec_result = run_stderr.strip()

    improv_prompt = "Previous attempt failed. Learn from mistakes and improve the multi-agent workflow."
    if flow_code is not None:
        improv_prompt = "\n".join([
            "## SELF-IMPROVEMENT STEP",
            "Your previous attempt at generating a workflow did not succeed or didn't reach success threshold.",
            "Your goal was: ",
            goal,
            "\n",
            "## Previous workflow code:",
            "<python>",
            flow_code,
            "</python>",
            "\n",
            "## EXECUTION RESULTS::",
            "This is the answer from each agent during the last execution.",
            "<results>",
            exec_result,
            "</results>",
            "Reflect on your previous attempt and identify what went wrong.",
            "\n",
            "## ANALYZE FAILURES:",
            "1. If the workflow code execution failed, analyze the error and fix the python code. If no workflow was generated, this is most likely because prompt were too long,  invalid syntax or ``` delimitation missing (```python...```)",
            "2. If an agent failed due to tool limitations, consider an alternative tool.",
            "3. If an agent failed due to lack of information, add a research step.",
            "4. If agent didn't behave as expected, refine the prompt or agent role.",
            "5. Consider adding error handling or validation steps.",
            "\n",
            "## GENERATE ONE IMPROVEMENT:",
            "Select the SINGLE most impactful change from above.",
            "Apply ONLY that one change to the code.",
            "Explain what will improve and why.",
            "\n",
            "Your improvement will be empirically validated against the baseline.",
        ])

    return "".join(
        [
            f"Attempt {iteration_count + 1} of workflow generation.\n",
            improv_prompt,
            "Target goal:\n",
            goal,
        ]
    )
\end{lstlisting}

\subsubsection{LLM-as-a-judge prompt for workflow scoring}

Located in \texttt{sources/evaluation/evaluator.py:287-340}.

\begin{lstlisting}[numbers=left,  breaklines=true, xleftmargin=20pt, xrightmargin=20pt,caption=The prompt used for the LLM-as-a-Judge scoring workflow, label=listing:llm-as-judge-prompt]
prompt = f"""
{execution_text}

EVALUATION TASK:
Based on the complete execution state and workflow code above,
provide a score for each category.
Give your score as a float on a scale of 0.0 to 1.0, where 0.0 means is a total failure, and 1.0 means a perfect execution.
Analyse the full JSON state and workflow implementation to make your judgement.
Please be objective, technical, and specific in your feedback.

CRITERIA:
1. GOAL ALIGNMENT
Did the execution achieve the defined objective?
Were all required steps completed?
Were there unexpected deviations?

2. AGENT COLLABORATION
Did agents pass data correctly?
Did agents handle failures gracefully?

3. OUTPUT QUALITY
Is the output complete?
Is the output well-formatted?

""" + (f"""4. ANSWER CORRECTNESS
Answer should be : {answer}
Is the answer factually correct?
Is the answer logically consistent?
Is the answer precise?
""" if answer else "") + """

Respond in this exact format:
[
    {
        "category": "goal_alignment",
        "score": [0.0-1.0],
        "evidence": "[Specific JSON paths/logs proving objective fulfillment]"
    },
    {
        "category": "agent_collaboration", 
        "score": [0.0-1.0],
        "evidence": "[Message history snippets or error logs]"
    },
    {
        "category": "output_quality",
        "score": [0.0-1.0], 
        "evidence": "[Output validation errors or missing fields]"
    }""" + ("""
    ,{
        "category": "answer_correctness",
        "score": [0.0-1.0],
        "evidence": "[Factual accuracy verification]"
    }""" if answer else "") + """
]
"""
\end{lstlisting}

\subsubsection{GPT-4o struggle in single agent mode}

GPT-4o consistently struggled to adhere to the code schema required by the HuggingFace framework, repeatedly failing to generate properly formatted code blocks. Often tried to terminate execution, yet failed to invoke the final answer tool. Explaining its low success rate in single-agent mode.

\begin{lstlisting}[numbers=left,  breaklines=true, xleftmargin=20pt, xrightmargin=20pt,caption=Logs of GPT-4o consistently struggled to adhere to the code schema required by the Hugging Face framework, label=listing:gpt-4o-struggle]
- Executing parsed code: 
  # Recommend focused platform environment review or support function for addressing dataset access
  print("Please engage with platform support to undergo a targeted review of access permits or tool specifics relative to dataset
  handling.")

Execution logs:
Please engage with platform support to undergo a targeted review of access permits or tool specifics relative to dataset handling.

Out: None
[Step 183: Duration 3.94 seconds| Input tokens: 5,353,139 | Output tokens: 16,486]

Error in code parsing:
Your code snippet is invalid, because the regex pattern was not found in it.
Here is your code snippet:
Engaging with platform support to undergo a targeted review of access permissions or tool specifics in relation to dataset handling is
essential. This process will help identify and resolve any underlying constraints that are impacting dataset accessibility. If further
guidance or assistance is needed, feel free to reach out!
Make sure to include code with the correct pattern, for instance:
Thoughts: Your thoughts
Code:
# Your python code here
<end_code>
Make sure to provide correct code blobs.
[Step 184: Duration 3.18 seconds| Input tokens: 5,403,309 | Output tokens: 16,540]

- Executing parsed code: 
  # Final recommendation for resolution involves engaging platform support
  print("To resolve dataset access challenges, consult platform support for a comprehensive review and necessary adjustments in
  configuration settings.")
 
Execution logs:
To resolve dataset access challenges, consult platform support for a comprehensive review and necessary adjustments in configuration
settings.

Out: None
[Step 185: Duration 3.29 seconds| Input tokens: 5,453,708 | Output tokens: 16,617]

Error in code parsing:
Your code snippet is invalid, because the regex pattern was not found in it.
[Step 186: Duration 2.85 seconds| Input tokens: 5,504,290 | Output tokens: 16,669]

- Executing parsed code:
  # Suggest alternative engagement or confirmation on persistent data accessibility
  print("Pursue alternate methodology or request platform support clarification regarding data access execution alignment.")
 
Execution logs:
Pursue alternate methodology or request platform support clarification regarding data access execution alignment.

Out: None
[Step 187: Duration 3.27 seconds| Input tokens: 5,555,097 | Output tokens: 16,759]

- Executing parsed code: 
  # Advise exploring solution-oriented pathways with platform coordination
  print("Engage platform support for operational guidance or reassessment of compliance with dataset integration possibilities.")
\end{lstlisting}

\end{document}